%% file: janus_tbiom_ieee.tex
\documentclass[10pt,journal,compsoc]{IEEEtran}

\usepackage{times,amsmath,epsfig,algorithm,amsfonts,amssymb,theorem,subfigure,cite,wrapfig,subfigure,multirow,adjustbox}
\usepackage[noend]{algorithmic}

 \makeatletter
\newcommand\figcaption{\def\@captype{figure}\caption}
\newcommand\tabcaption{\def\@captype{table}\caption}

\graphicspath{{figs/}}

\usepackage{cite}
\begin{document}
\title{A Fast and Accurate System for Face Detection, Identification, and Verification}
%
%
%
%

\author{Rajeev Ranjan, Ankan Bansal, Jingxiao Zheng, Hongyu Xu, Joshua Gleason, Boyu Lu, Anirudh
    Nanduri, Jun-Cheng Chen, Carlos D. Castillo, Rama Chellappa
\IEEEcompsocitemizethanks{\IEEEcompsocthanksitem University of Maryland, College Park, MD.\protect\\
E-mail: ankan@umiacs.umd.edu}
\thanks{Manuscript received ; revised }}

%
%

\markboth{Journal of \LaTeX\ Class Files,~Vol.~14, No.~8, August~2015}%
{Shell \MakeLowercase{\textit{et al.}}: Bare Demo of IEEEtran.cls for Computer Society Journals}
%



\IEEEtitleabstractindextext{%
\input{abstract}

\begin{IEEEkeywords}
    Face recognition, Face identification/verification, Face detection, Deep Learning
\end{IEEEkeywords}}

\maketitle

\IEEEdisplaynontitleabstractindextext

%
\IEEEpeerreviewmaketitle

\input{intro}

\input{related_work}
\input{approach}
\input{experiments}
\input{conclusion}


%
%
\bibliographystyle{IEEEtran}
\bibliography{all,dataset_refs,sps_refs,detection_refs,attribute}

%

\begin{IEEEbiographynophoto}{Rajeev Ranjan} received the B.Tech. degree in Electronics and
    Electrical Communication Engineering from Indian Institute of Technology Kharagpur, India, in
    2012. He is currently a Research Assistant at University of Maryland College Park. His research
    interests include face detection, face recognition and machine learning. He received Best Poster
    Award at IEEE BTAS 2015. He is a two-time recipient of UMD Outstanding Invention of the Year
    award in the area of Information Science. He received the 2016 Jimmy Lin Award for Invention.
\end{IEEEbiographynophoto}

\begin{IEEEbiographynophoto}{Ankan Bansal} (B.Tech-M.Tech, IIT Kanpur, 2015) is a PhD student at the
    University of Maryland, College Park. His research interests include multi-modal learning,
    action understanding, and face analysis. He was awarded the Clark School of Engineering
    Distinguished Graduate Fellowship, 2015-2016.
\end{IEEEbiographynophoto}

\begin{IEEEbiographynophoto}{Jingxiao Zheng} (B.Eng, SEU, Nanjing, 2012, M.S., UMD,
    2014) is currently a research assistant in the Institute for Advanced Computer Studies at the
    University of Maryland, College Park. His advisor is Prof. Rama Chellappa. His research
    interests include deep learning, face recognition, and action recognition. 
\end{IEEEbiographynophoto}

\begin{IEEEbiographynophoto}{Hongyu Xu} (B.Eng, USTC, 2012, M.S., UMD, 2016) is currently a research
    assistant in the Institute for Advanced Computer Studies at the University of Maryland, College
    Park, advised by Prof. Rama Chellappa. He is a former research intern with Snap Research
    (summer, fall 2017) and Palo Alto Research Center (PARC) (summer 2014). His research interests
    include object detection, deep learning, dictionary learning, face recognition, object
    classification, and domain adaptation.
\end{IEEEbiographynophoto}

\begin{IEEEbiographynophoto}{Joshua Gleason} (B.S. Elec. Eng., B.S. Comp. Sci., UNR, 2013) is a PhD
    student at the University of Maryland, College Park and is advised by Prof. Rama Chellappa. His
    current research interests include face recognition, transfer learning, and general machine
    learning and computer vision topics.
\end{IEEEbiographynophoto}

\begin{IEEEbiographynophoto}{Boyu Lu} (B.Eng, USTC, 2012) is currently a research assistant at the
    University of Maryland, College Park. His research interests include face recognition, deep
    learning, and domain adaptation.
\end{IEEEbiographynophoto}

\begin{IEEEbiographynophoto}{Anirudh Nanduri} (B.Tech, IIT Bombay, 2015) is a PhD student at the
    University of Maryland, College Park. His research interests include geometric deep learning,
    medical image processing, face analysis and action recognition. He was awarded the Clark School
    of Engineering Distinguished Graduate Fellowship, 2015-2016.
\end{IEEEbiographynophoto}

\begin{IEEEbiographynophoto}{Jun-Chen Chen} (Ph.D., UMD, 2016) is a postdoctoral research fellow at
    the University of Maryland Institute for Advanced Computer Studies (UMIACS). His current
    research interests include computer vision and machine learning with applications to face
    recognition and facial analysis. He was a recipient of ACM Multimedia best technical full paper
    award, 2006.
\end{IEEEbiographynophoto}


\begin{IEEEbiographynophoto}{Carlos D. Castillo} (Ph.D., UMD, 2012) is an assistant research
    scientist at the University of Maryland Institute for Advanced Computer Studies (UMIACS). His
    current research interests include stereo matching, multi-view geometry, face detection,
    alignment and recognition.
\end{IEEEbiographynophoto}

\begin{IEEEbiographynophoto}{Rama Chellappa} (Ph.D., Purdue University, 1981) is a Distinguished University
    Professor, and a Minta Martin Professor of Engineering of Electrical and Computer
    Engineering Department at University of Maryland, College Park. His current research interests
    are face and gait analysis, 3-D modeling from video, image and video-based recognition and
    exploitation, compressive sensing, and hyper spectral processing. He received the Society,
    Technical Achievement and Meritorious Service Awards from the IEEE Signal Processing Society.
\end{IEEEbiographynophoto}




\end{document}

%% file: abstract.tex
\begin{abstract}
    The availability of large annotated datasets and affordable computation power have led to impressive
    improvements in the performance of CNNs on various object detection and recognition benchmarks.
    These, along with a better understanding of deep learning methods, have also led to improved
    capabilities of machine understanding of faces. CNNs are able to detect faces, locate facial
    landmarks, estimate pose, and recognize faces in unconstrained images and videos. In this paper,
    we describe the
    details of a deep learning pipeline for unconstrained face identification and verification which achieves
    state-of-the-art performance on several benchmark datasets. We propose a novel face detector,
    Deep Pyramid Single Shot Face Detector (DPSSD), which is fast and capable of detecting faces
    with large scale variations (especially tiny faces).
    We give design details of the various
    modules involved in automatic face recognition: face detection, landmark localization and alignment, and
    face identification/verification. We provide evaluation results of the proposed face detector on challenging
    unconstrained face detection datasets. Then, we present experimental results for IARPA Janus Benchmarks A, B and C
    (IJB-A, IJB-B, IJB-C), and the Janus Challenge Set 5 (CS5). 
\end{abstract}

%% file: intro.tex
\section{Introduction}\label{sec:intro}

Facial analytics is an active area of research. It involves extracting information
such as landmarks, pose, expression, gender, age, identity etc. It has several application including
law enforcement, active authentication on devices, face biometrics for payments, self-driving
vehicles etc.

Face identification and verification systems typically have three modules. First, a face detector for
localizing faces in an image is needed. Desirable properties of a face detector are robustness to variations
in pose, illumination, and scale. Also, a good face detector should be able to output consistent and
well localized bounding boxes. The second module localizes the facial
landmarks such as eye centers, tip of the nose, corners of the mouth, tips of ear lobes,
          etc. These landmarks are used to align faces which mitigates the effects of in-plane
          rotation and
          scaling. Third, a feature extractor encodes the identity information in a high-dimension
          descriptor.
          These descriptors are then used to compute a similarity score between two faces. An
          effective feature
          extractor needs to be robust to errors introduced by previous steps in the pipeline:
          face
          detection, landmark localization, and face alignment. 

          CNNs have been shown to be very effective for several computer vision tasks like image
          classification \cite{krizhevsky_imagenet_2012, szegedy_going_2014, he2016deep}, and object
          detection
          \cite{girshick_rich_2014, ren2015faster, liu2016ssd}. Deep CNNs (DCNNs) are highly non-linear
          regressors
          because of the presence of hierarchical convolutional layers with non-linear activations.
          DCNNs have
          been used as building blocks for all three modules of automatic face recognition: face detection
          \cite{najibi2017ssh,ranjan2016all,ranjan2015deep,ranjan2017hyperface}, facial keypoint
          localization \cite{kumar_face_2016, ranjan2016all,ranjan2017hyperface}, and face
          verification/identification
          \cite{chen2016end, bansal2018deep} Ever-increasing computation power and availability of
          large datasets like CASIA-WebFace
          \cite{yi_learning_2014}, UMDFaces \cite{bansal2017s, UMDFaces}, MegaFace
          \cite{kemelmacher2016megaface, nech2017level}, MS-Celeb-1M \cite{guo2016ms}, VGGFace
          \cite{parkhi_deep_2015, vggface2}, and WIDER Face \cite{yang2016wider} has led to
          significant
          performance gains from DCNNs. This is because of the large variations in pose,
          illumination, and
          scale of faces present in these datasets.

          This paper makes two primary contributions: 1) We propose a novel face detector that is
          fast and can detect faces over a large variation of scale. 2) Present a
          DCNN-based automatic face
          recognition pipeline which achieves impressive results on several recent benchmarks. We
          require our face detector to be both fast and accurate in order to build an efficient
          end-to-end face recognition pipeline. Hence, we design a face detector that provides the
          output in a single pass of the network. In order to detect faces at different scales, we
          make use of the inbuilt pyramidal hierarchy present in a DCNN, instead of creating an
          image pyramid. This further reduces the processing time. We develop specific anchor filters
          for detecting tiny faces. We apply the bottom-up approach to incorporate contextual
          information, by adding features from deeper layers to the features from shallower
          layers. The proposed face detector is called Deep Pyramid Single Shot Face Detector
          (DPSSD).

          Once we get the face detections from DPSSD, we follow the pipeline to localize landmarks
          and extract deep identity features for face recognition and verification. Each module of
          the presented recognition pipeline (face detection, landmark localization, and feature
                  extraction) is based on DCNN models. We use an
          ensemble of CNNs as feature extractors and combine the features from the DCNNs into the
          final
          feature representation for a face. In order to localize facial landmarks for
          face alignment, we use the DCNN architecture proposed in \cite{ranjan2016all}. We describe
    each of the modules in detail and discuss their performance on the challenging IJB-A, IJB-B,
    IJB-C (see figure \ref{fig:ijbc_example} for a sample of faces), and IARPA Janus Challenge Set 5 (CS5)
    datasets. We also present an overview of recent approaches in this area, and discuss their
    advantages and disadvantages.

    The automatic face verification pipeline presented in this paper is significantly improved from
    its
    predecessors \cite{chen2016end, ranjan2018deep} in the following ways:
    (1) It uses a much more effective Crystal Loss function described in \cite{ranjan2018crystal} to
    train the
    networks. Crystal Loss creates more concentrated clusters of classes and increases inter-class
distances. (2) It employs an efficient metric learning method (Triplet Probabilistic Embedding)
    \cite{swami_btas_2016}. TPE uses inner-product based constraints instead of the commonly
    used
    norm-based constraints while optimizing the embedding matrix. 

    The paper is organized as follows. First we discuss recent developments in face
    detection (section~\ref{sec:face_detection}), keypoint detection (section
            \ref{sec:fid_detection}), face
    recognition (section \ref{sec:face_verification}) and multi-task learning
    (MTL)~\ref{sec:mtl_pipeline}. Then, in section \ref{sec:approach}, we describe our
    pipeline
    for face detection, identification, verification, and recognition and present results in section
    \ref{sec:experiments}. We discuss some open issues and conclude in section \ref{sec:conclusion}.


%% file: related_work.tex
\section{A Brief Survey of Existing Literature}\label{sec:related_work}

We give a brief overview of recent works on different modules of a face identification/verification
pipeline.
We first discuss recent face detection methods. Then we consider the second module: facial
keypoint detection. Finally, we discuss several recent works on feature learning and
summarize much of the state-of-the-art work on face verification and identification.

\input{detection.tex}
\input{verification.tex}


\subsection{Multi-Task Learning for Facial Analysis} 
\label{sec:mtl_pipeline}

Multi-Task Leaning is the setting where multiple parts of a problem are tackled simultaneously,
    usually using the same features. The idea behind MTL learning is that different tasks can
    benefit
    from each other. The MTL framework was first used and analyzed by Caruana
    \cite{caruana1998multitask}.
    Zhu \emph{et al.} \cite{AFW_dataset_CVPR2012} proposed a multi-task approach for simultaneous
    face
    detection, landmark localization, and head-pose estimation. MTL has been shown to improve the
    performance for the tasks involved by leveraging information from different supervision sources.
    For
    example, JointCascade \cite{JointCascade_LI_ECCV2014} achieved improvement in face detection
    performance by adding landmark localization to face detection during training. 

    However, because the above mentioned methods used hand-crafted features, extending these
    to
    new tasks is difficult. Different tasks required different types of specialized hand-crafted
    features. For
    example, face detection usually used Histograms of Oriented Gradients (HOG), whereas face
    recognition typically used Local Binary Patterns (LBP). Combining these to achieve concurrent
    face
    detection and recognition is difficult. However, features obtained from DCNNs can encode various
    properties of the visual data. Contrary to hand-designed
    features, it is possible to train a single DCNN which can accomplish multiple tasks such as face
    detection, landmark localization, attribute prediction, age estimation, face recognition etc. at
    the
    same time. Shared deep features help in exploiting the relationship between different tasks.
    Using
    MTL can be considered an additional regularization for the CNN
    \cite{Goodfellow-et-al-2016-Book}.

    HyperFace \cite{ranjan2017hyperface} is among the first few multi-task methods for face
    analysis.
    It was designed for simultaneous face detection, keypoint localization, head-pose estimation,
    and
    gender classification. It exploited the synergy among various tasks by sharing 
     location-specific features from lower layers of a CNN and semantically rich
    features
    from higher layers. This helped in improving the performance for each task. Similarly, TCDCN
    \cite{TCDCN} added head yaw estimation, gender recognition, smile and glass detection to the
    task of
    landmark localization. These auxiliary tasks improved the performance of landmark
    localization. The All-in-One Face \cite{ranjan2016all} network extended HyperFace by adding more tasks and
    training data. Our approach uses All-in-One Face for facial keypoint detection and face
    alignment. We give a brief overview in section \ref{sec:aio}.  Table \ref{tbl:mtl} summarizes
    the tasks performed by some recent MTL face analysis methods. 

    \begin{table*}[htp!]
    \centering
    \resizebox{\textwidth}{!}
{
    \begin{tabular}{|l|l|l|l|l|l|l|l|l|}
    \hline
        Method          & Face Detection            & Fiducials
        & Head-Pose  & Gender & Age & Expression & Other
        Attributes & Face Recognition\\
        \hline\hline
        Zhu~\emph{et
            al.}~\cite{AFW_dataset_CVPR2012}
    & \checkmark  & \checkmark  & \checkmark    &    &    &  & &\\
        \hline
        JointCascade~\cite{JointCascade_LI_ECCV2014}
    & \checkmark  & \checkmark  &  &    &    &  & &\\
        \hline
        Zhang~\emph{et al.}~\cite{zhang2016joint}
    & \checkmark  & \checkmark  &  &    &    &  & &\\
        \hline
        TCDCN~\cite{TCDCN}        &   & \checkmark  & \checkmark
        & \checkmark   &    & \checkmark  & \checkmark &\\ 
        \hline
        HyperFace~\cite{ranjan2017hyperface}        &
        \checkmark  & \checkmark  & \checkmark &
        \checkmark   &    &   & &\\ 
        \hline
        He~\emph{et al.}~\cite{he2017jointly} &
        \checkmark  &   &  & \checkmark   &
        \checkmark   & \checkmark  & \checkmark
        &\\  
        \hline
        DAGER~\cite{dehghan2017dager} &
        &   &  & \checkmark   &
        \checkmark   & \checkmark  & &
        \\
        \hline
        All-In-One
        Face~\cite{ranjan2016all}
    & \checkmark  & \checkmark  & \checkmark & \checkmark   & \checkmark
        & \checkmark  & & \checkmark\\ 
        \hline
        \end{tabular}
}
\vspace{2mm}
\caption{List of various MTL-based facial analysis algorithms along with the types of face tasks
    they can perform }
    \label{tbl:mtl}

    \end{table*}

%% file: detection.tex
\subsection{Face Detection}
\label{sec:face_detection}

Face detection is the first step in any face recognition/verification pipeline. A face detection
algorithm outputs the locations of all faces in a given input image, usually in the form of bounding
boxes. A face detector needs to be robust to variations in pose, illumination, view-point,
expression, scale, skin-color, some occlusions, disguises, make-up, etc. Most recent DCNN-based face
detectors are inspired by general object detection approaches. CNN detectors can be divided into
two sub-categories: 1) region-based, and 2) sliding window-based.

\textbf{Region-based} approaches first generate a set object-proposals and use a CNN classifier to classify
each proposal as a face or not face. The first step is usually an off-the-shelf proposal generator
like selective search \cite{uijlings2013selective}. Some recent detectors which use this approach
are HyperFace \cite{ranjan2017hyperface}, and All-in-One Face \cite{ranjan2016all}. Instead of
generating object proposals by a generic method, Faster R-CNN \cite{ren2015faster} used a Region
Proposal Network (RPN). Jiang and Learned-Miller used a Faster R-CNN network to detect faces in
\cite{jiang2016face}. Similarly, \cite{li2016face} proposed a multi-task face detector based on the
Faster-RCNN framework. Chen \emph{et al.} \cite{chen2016supervised} trained a multi-task
RPN for face detection and facial keypoint localization. This allowed them to reduce redundant
face proposals and improve the quality of face proposals. The Single Stage Headless face detector
\cite{najibi2017ssh} is also based on an RPN. 

\textbf{Sliding window-based} methods output face detections at every location in a feature map at a given
scale. These detections are composed of a face detection score and a bounding box. This approach
does not rely on a separate proposal generation step and is, thus, much faster than region-based
approaches. In some methods \cite{ranjan2015deep, farfade2015multi}, multi-scale detection is
accomplished by creating an image pyramid at multiple scales. Similarly, Li \emph{et al.}
\cite{li2015convolutional} used a cascade architecture for multiple resolutions. The Single Shot
Detector (SSD) \cite{liu2016ssd} is also a multi-scale sliding-window based object detector.
However, instead of using an object pyramid for multi-scale processing, it utilizes the hierarchal
nature of deep CNNs. Methods like ScaleFace \cite{yang2017face}, and S3FD \cite{zhang2017s} use
similar techniques for face detection.



In addition to the development of improved detection algorithms, rapid progress in face detection
performance has been spurred by the availability of large annotated datasets. FDDB \cite{fddbtech}
consists of $2,845$ images containing a total of $5,171$ faces. 
Similar in scale is the MALF
\cite{yang2015fine} dataset which contains $5,250$ images with $11,931$ faces. A much larger dataset
is WIDER Face \cite{yang2016wider}. It contains over $32,000$ images containing faces with large
variations in expression, scale, pose, illumination, etc. Most state-of-the-art face detectors have
been trained on the WIDER Face dataset. This dataset contains many tiny faces. Several of the above
mentioned face detectors still struggle with finding these small faces in images. Hu \emph{el al.}
\cite{hu2016finding} showed that context is important for detecting such faces.

An extensive survey of face detection methods developed before 2014 can be found in
\cite{zafeiriou2015survey}. Chen
\emph{et al.} \cite{chen2016end} discuss the importance of face association for face recognition in
videos. Association is the process of finding the correspondences between different faces in
different video frames. 


\subsection{ Facial Keypoints Detection and Head Orientation}
\label{sec:fid_detection}
Facial keypoints include corners of the eyes, nose tip, ear lobes, mouth corners etc. These are
needed for face alignment which is important for face identification/verification \cite{bansal2017s}. Head pose is
another important information of interest. A comprehensive survey of keypoint localization methods can be
found in \cite{wang2017facial} and \cite{chrysos2016comprehensive}. 


Facial keypoint detection methods can be divided into two types: model-based and regression-based.
The model-based approaches
create a representation of shape during training and use this to fit faces during testing. Model-based methods
include PIFA \cite{jourabloo2015pose}, and 3DDFA \cite{zhu2016face}. Jourabloo \emph{et al.}
\cite{jourabloo2016large} considered face alignment as a dense 3D model fitting problem and used a
cascade of DCNN-based regressors to estimate the camera projection matrix and 3D shape parameters.
Antonakos \emph{et al.} \cite{antonakos2015active} modeled appearances using multiple graph-based
pairwise normal distributions between patches.

Cascade regression-based methods directly map image appearance to the
target output. Zhang \emph{et al.} \cite{zhang2014coarse} used a cascade of several successive
stacked auto-encoder networks. This approach refines the coarse locations obtained from the first few
stacked auto-encoder networks using subsequent networks. Bulat \emph{et al.} also first roughly
localized each facial landmark and then refined the detection results. Similarly, the approach proposed by Sun
\emph{et al.} \cite{sun_deep_2013} fused outputs from multiple networks at each level of a cascade.
Another method which combined outputs from multiple regressors is cascade compositional learning
(CCL) \cite{zhu2016unconstrained}. Kumar \emph{et al.} \cite{kumar2017kepler} proposed an iterative
method for keypoint
estimation and pose predication. The method proposed by Trigeorgis \emph{et al}
\cite{trigeorgis2016mnemonic} jointly trained a convolutional recurrent neural network architecture. 
In another work, Kumar \emph{et al.} \cite{kumar_face_2016} developed a single CNN for keypoint
localization. 

The 300 Faces In-the-Wild database (300W) \cite{sagonas2016300} is a benchmark for a fair comparison
of different facial detection methods. It combines and extends several previously available
datasets like LFPW, Helen, AFW, Ibug \cite{wang2017facial} and 600 test images. 


Some works have also used generic 3D face models for face alignment/frontalization
\cite{hassner2015effective}. However, the advantages of such methods are limited and their
performance can easily be improved upon by multi-task learning (MTL) approaches. 


%% file: verification.tex
\subsection{Face Identification and
Verification}\label{sec:face_verification}

In this section, we provide a brief introduction to recent works on CNN-based face identification and
verification. Interested readers are referred to \cite{learned2016labeled} for a summary of methods
developed before the wide adoption of CNNs. 

A face identification/verification system has two main parts: 1) robust face
representation; and 2) a classifier (in case of identification) or similarity measure (for
verification). 

\subsubsection{Robust Face Representations}
Deep networks are able to learn discriminative features when trained with large datasets. Huang
\emph{et al.} \cite{huang2012learning} used convolutional deep belief networks based on local
restricted Boltzmann machines to learn face representations. Their models achieved good performance
on the LFW dataset without requiring large annotated face datasets. 

On the other hand, Taigman \emph{et al.} used a proprietary face dataset consisting of four million
faces of over 4,000 identities to train a nine-layer deep network (DeepFace) \cite{taigman_deepface_2014}.
Instead of using standard convolutional layers, they used several locally connected layers without
weight sharing. Similarly, FaceNet \cite{schroff_facenet_2015} was trained on a dataset of
about 200 million images of about 8 million identities. It directly optimized the embedding itself
using triplets of roughly aligned matching/non-matching face patches.

The DeepID frameworks \cite{sun2014deep, sun_deep_2014, sun_deeply_2014} utilized an ensemble of
smaller deep convolutional networks than DeepFace or FaceNet. Each DCNN consisted of four convolutional layers
and was trained with about 200,000 images of about 10,000 identities. Using an ensemble of models
and a large number of distinct identities helped DeepID learn discriminative face representations
which allowed it to achieve super-human face verification performance on the LFW dataset.

The CASIA-WebFace dataset \cite{yi_learning_2014} which consists of about 0.5 million face images from
10,575 subjects was used to train a DCNN with 5 million parameters. The model achieved satisfactory
performance and the dataset is widely used for training CNNs. Other large-scale datasets have
followed, \emph{e.g.} VGGFace \cite{parkhi_deep_2015}, VGGFace2 \cite{vggface2}, UMDFaces
\cite{UMDFaces, bansal2017s} etc. 

Parkhi \emph{et al.} \cite{parkhi_deep_2015} trained a CNN based on VGGNet \cite{simonyan2014very}
for face verification using the VGGFace dataset. This model achieved competitive results on both LFW
\cite{huang_labeled_2008} and YTF \cite{YTF} datasets.

Larger datasets and more difficult evaluation metrics require representations invariant to pose,
age, illumination etc. AbdAlmageed \emph{et al.} \cite{AbdAlmageed_pose_2016} trained separate DCNN
models for frontal, half-profile, and fill-profile faces as a way to handle pose variation. Adding
more images of faces in profile to the training set is another way of thinking about robustness. Masi
\emph{et al.} used 3D morphable models to augment the CASIA-WebFace dataset.  This has the added
advantage of not requiring large-scale human annotation efforts.

The most widely used softmax-loss usually does not lead to concentrated clustering of face
representations. Several modifications and replacements have been proposed to achieve enhanced
representations of faces. Ding \emph{et al.} \cite{ding2016trunk} proposed a new triplet loss
function which achieved state-of-the-art performance for video-based face recognition.
Wen \emph{et al.} \cite{wen2016discriminative} added a regularization constraint to the softmax loss
based on the centroid for each class. Liu \emph{et al.} \cite{liu2017sphereface} proposed angular
loss based on modified softmax. This led to a discriminative face representation which is optimized
for the most commonly used similarity metric, \emph{viz.}, cosine similarity. Ranjan \emph{et al.}
\cite{ranjan2017l2} regularized the softmax loss with a scaled $L_2$-norm constraint. This achieved
state-of-the-art results on IJB-A \cite{IJBA}. 

Video-face recognition and template-based face processing requires using feature aggregation methods
which combine features from several face images into one. Yang \emph{et al.} \cite{yang_neural_2016}
proposed a dynamically weighted aggregation approach (Neural Aggregation Network). Similarly, Bodla
\emph{et al.} \cite{bodla2017deep} used a neural network to fuse facial features from two different
DCNN models. However, as we show later, a simple average aggregation strategy appears to be
sufficient to equal the performances of these methods.

\subsubsection{Discriminative Metric Learning}
Learning a classifier or a similarity metric is the next step in obtaining robust facial features.
For face verification, features for two faces belonging to the same person should be similar while
features for face belonging to different persons should be dissimilar. Several recent works have
come up with ways of encoding this requirement in the training loss functions or network designs.

The first approach uses pairs of images to train a feature embedding where positive pairs are closer
and negative pairs are farther apart. Hu \emph{et al.} \cite{hu_discriminative_2014} used deep
neural networks to learn a discriminative metric. Schroff \emph{et al.} \cite{schroff_facenet_2015},
Parkhi \emph{et al.} \cite{parkhi_deep_2015}, and Swami \emph{et al.} \cite{swami_btas_2016} used a
triplet loss to embed DCNN features into a discriminative subspace. This led to performance
improvements on face verification.

Another approach is to modify the commonly used cross-entropy loss to incorporate the discriminative
constraint. Wen \emph{et al.} \cite{wen2016discriminative} introduced the center loss for learning
discriminative face embeddings. Ranjan \emph{et al.} presented crystal loss \cite{ranjan2018crystal}
which uses a feature normalization and scaling before the softmax loss. Similarly DeepVisage
\cite{hasnat2017deepvisage} normalized the features using a special case of batch normalization.
SphereFace \cite{liu2017sphereface} proposed angular softmax which yields angularly discriminative
features. CosFace \cite{wang2018cosface} $L_2$ normalizes both features and weights to
remove radial variations and introduces a cosine margin term to maximize the decision margin in
angular space.

\subsubsection{Implementation}
Obtaining discriminative and robust features is important for both face verification and
identification. For face verification, given a pair of faces, the two face features are compared
using a similarity metric. $L_{2}$ distance and cosine similarity are the two most commonly used
metrics for comparing two face feature representations. For identification, the feature of a given
probe face is compared against a large gallery and the most similar gallery faces give the identity
of the probe face. To obtain robust features, ensemble of DCNNs can be used to extract different
face representations which can be later fused into a single robust representation \cite{sun2014deep,
sun_deep_2014, sun_deeply_2014, bodla2017deep}. 


Deep networks are extremely data hungry. There are several publicly available face datasets which
can be used to train deep networks for face identification and verification. Table \ref{tbl:ijba}
presents the details of some of these datasets. 

\begin{table}[H]
	\begin{center}
		\tabcolsep=0.15cm 		\begin{adjustbox}{max width=\textwidth}
			\begin{tabular}{|c|c|c|}
				\hline
				\multicolumn{3}{|c|}{Face Recognition}\\
				\hline
				Name & \#faces & \#subjects\\
				\hline
				MS-Celeb-1M~\cite{guo2016ms} & 10M & 100K\\
				CelebA~\cite{liu2015deep} & 202,599 & 10,177\\
				CASIA-WebFace~\cite{yi_learning_2014} & 494,414 & 10,575\\
				VGGFace~\cite{parkhi_deep_2015} & 2.6M & 2,622\\
				Megaface~\cite{kemelmacher2016megaface,nech2017level} & 4.7M & 672K\\
				LFW~\cite{huang_labeled_2008} & 13,233 & 5749\\
				IJB-A~\cite{IJBA} & 5,712 images, 2,085 videos & 500\\
				IJB-B~\cite{IJBB} & 11,754 images, 7,011 videos & 1,845\\
				IJB-C~\cite{IJBC} & 31,334 images, 11,779 videos & 3,531\\
				YTF~\cite{YTF} & 3,425 videos & 1,595\\
				PaSC~\cite{beveridge2013challenge} & 2,802 videos & 293\\
				CFP~\cite{sengupta2016frontal} & 7,000 & 500 \\
				UMDFaces~\cite{UMDFaces} & 367,888 & 8,277\\
                UMDFace Video~\cite{bansal2017s} & 22,075 videos & 3,107\\
                VGGFaces2~\cite{vggface2} & 3.31M & 9,131 \\
				\hline
			\end{tabular}
		\end{adjustbox}
        \vspace{2mm}
		\caption{Recent datasets for face recognition.}
		\label{tbl:ijba}

	\end{center}
\end{table}

Pre-processing and model/dataset selection are extremely important decisions that need to be made
before training face recognition systems. Recently, Bansal \emph{et al.} \cite{bansal2017s}
studied the good and bad practices for such decisions. They tried to answer the following questions:
(1) Can we train on still images and expect the systems to work on videos? (2) Are deeper datasets
better than wider datasets where given a set of images deeper datasets mean more images per subject,
and wider datasets mean more subjects? (3) Does adding label noise always leads to improvement in
performance of deep networks? (4) Is alignment needed for face recognition? They \cite{bansal2017s}
essentially demonstrated the importance of using clean training data, good face alignment, and
training deep networks with a combination of still images and video frames.

\begin{figure*}[htp!]
\begin{center}
 \includegraphics[width=4in]{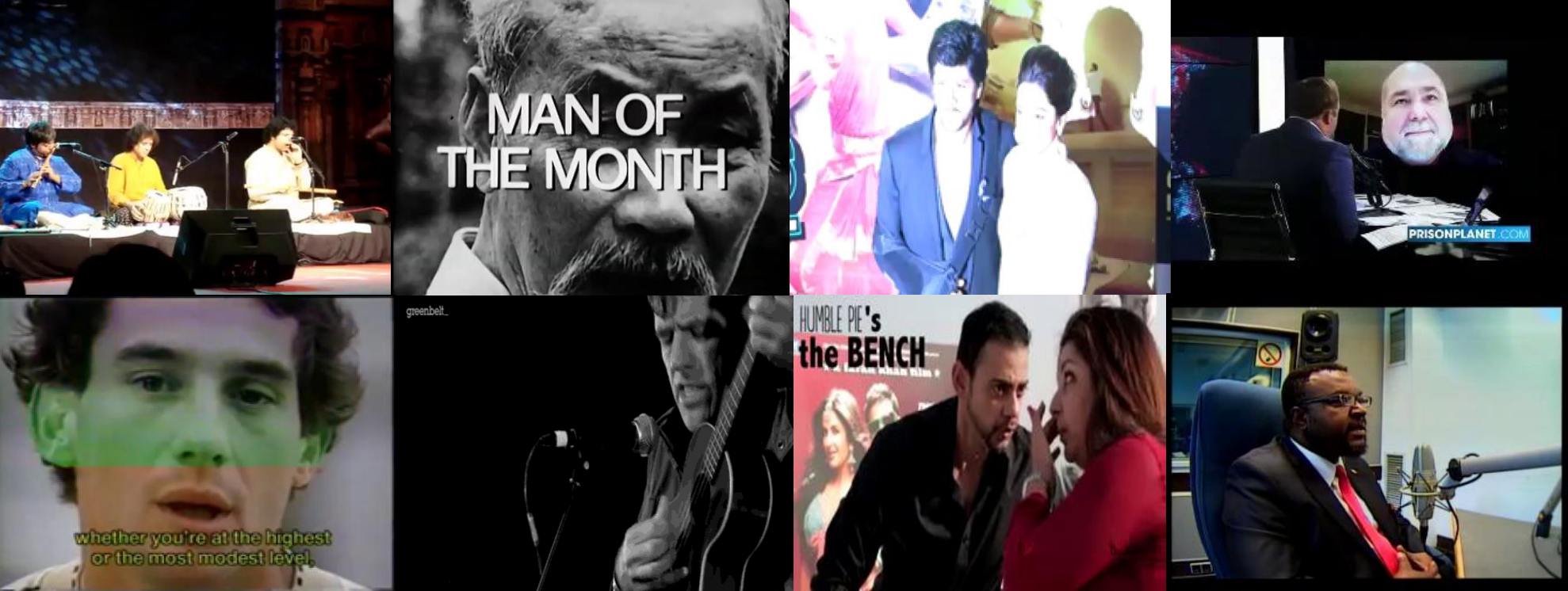}
\end{center}
  \caption{Some samples from the IJB-C dataset. This shows the wide range of image quality, pose,
      illumination, and expression variation in images.}
  \label{fig:ijbc_example}
\end{figure*}



%% file: approach.tex
\section{A State-of-the-art Face Verification and Recognition Pipeline}
\label{sec:approach}

\begin{figure*}[htp]
    \centering
    \includegraphics[height=3in]{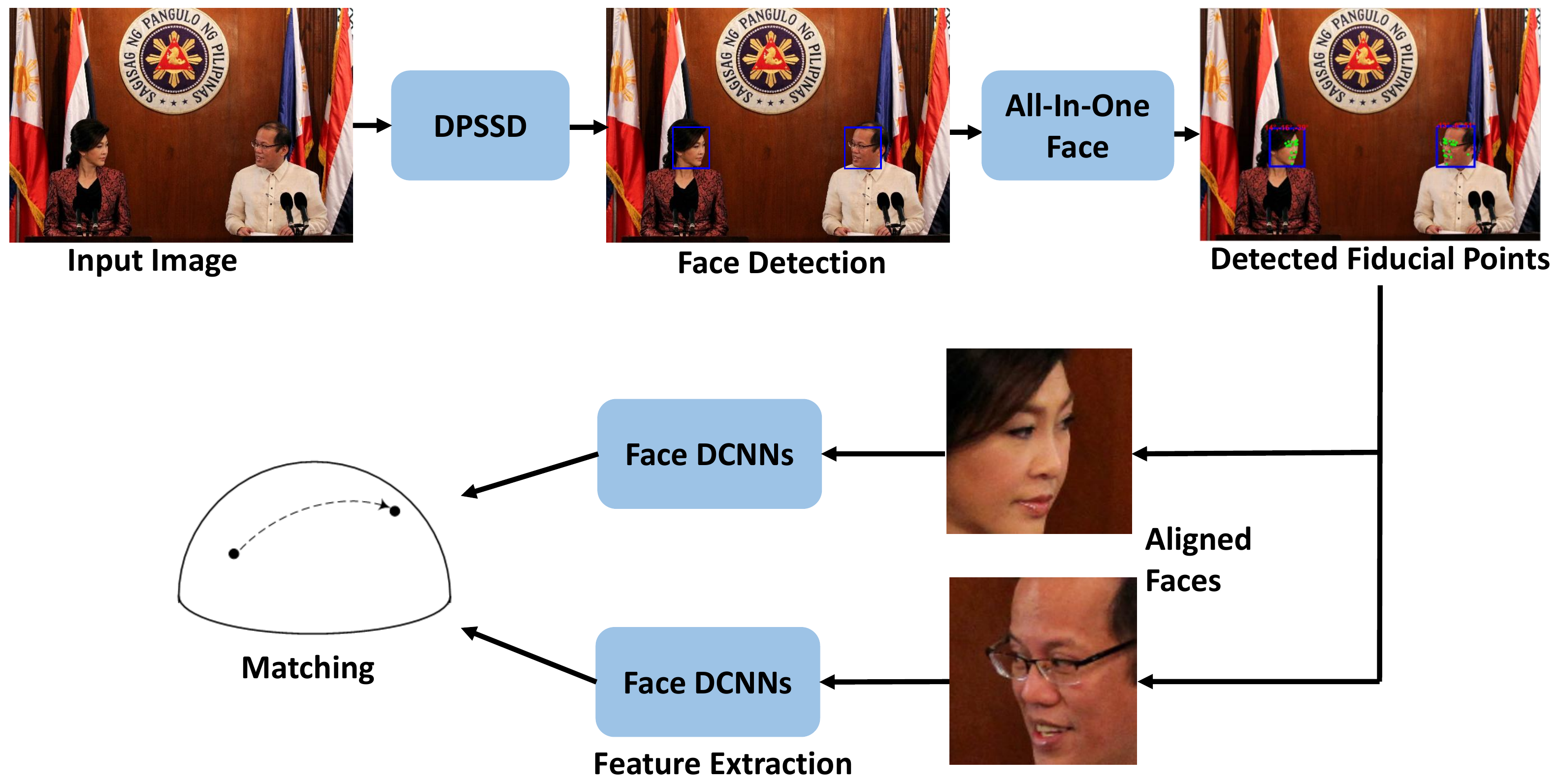}
    \caption{Our face recognition pipeline. We detect faces using our proposed DPSSD face detector
        (section \ref{sec:dpssd}).
        These detections are passed to the All-in-One Face network (section \ref{sec:aio}) which
        outputs facial keypoints for each face. These
are used to align faces to canonical views. We pass these aligned faces through our face
representation networks (section \ref{sec:face_id_ver}) and obtain the similarity between two faces
using cosine similarity.} 
    \label{fig:pipeline}
\end{figure*}

In this section, we discuss a state-of-the-art pipeline for face identification and verification, built
by authors over the last eighteen months. An overview of the pipeline is given in figure
\ref{fig:pipeline}. We first introduce the proposed DPSSD face detector in
subsection~\ref{sec:dpssd}. We then briefly summarize our face alignment method using the MTL approach.
Lastly, we describe our approach for extracting identity features and using them for face
identification and verification.

\input{dpssd.tex}

\subsection{Face Alignment using All-In-One Face}
\label{sec:aio}

\begin{figure}[htp]
    \centering
    \includegraphics[height=3in]{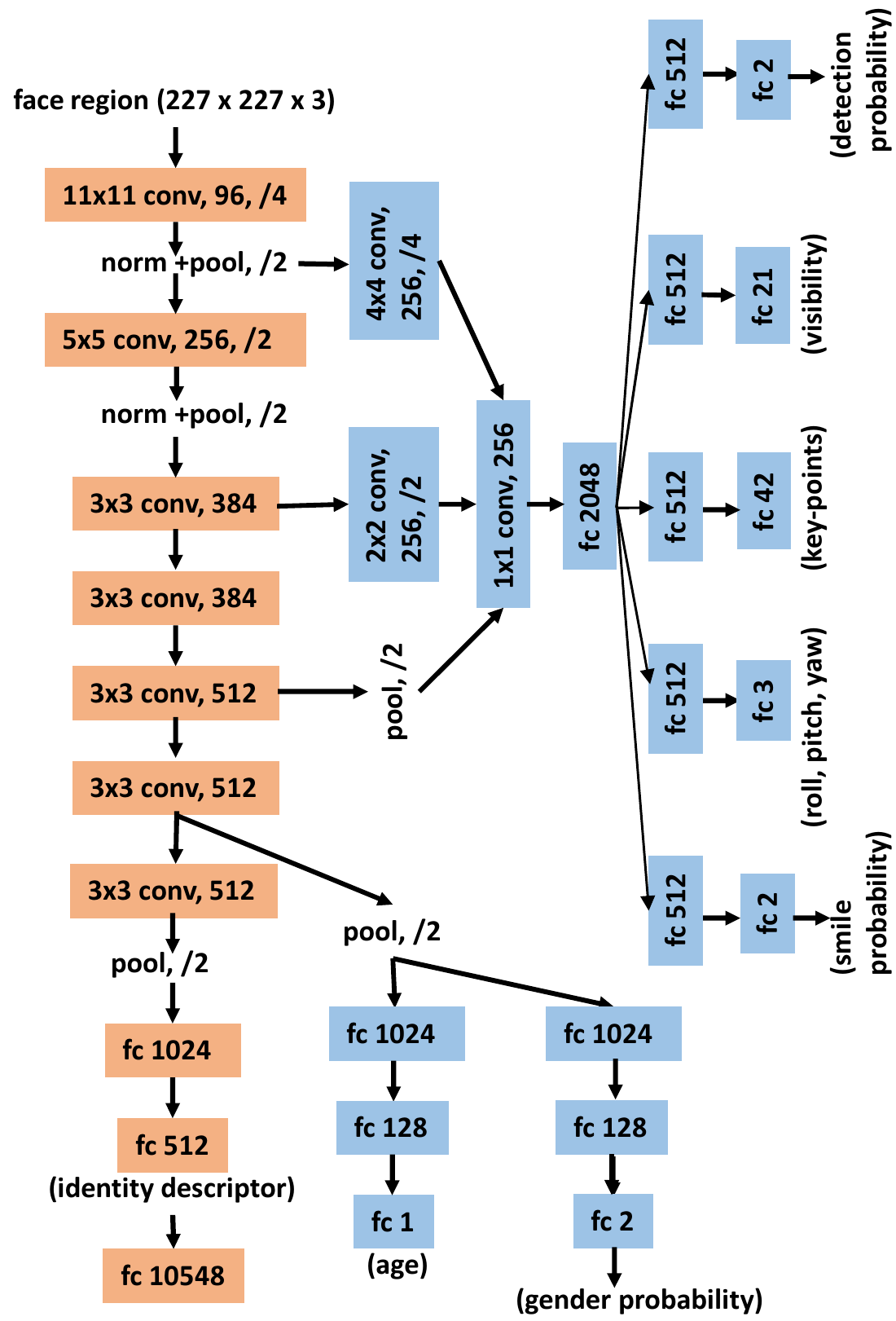}
	\caption{The All-In-One Face network architecture.} 
    \label{fig:system_overview}
\end{figure}


The proposed system for face identification and verification uses the All-in-One Face framework
\cite{ranjan2016all} for keypoint localization. The All-In-One Face~\cite{ranjan2016all} is a recent
method that simultaneously performs the tasks of face detection, landmarks localization, head-pose
estimations, smile and gender classification, age estimation and face recognition and verification.
The model is trained jointly for all these tasks in a MTL framework, which builds up a synergy that
helps in improving the performance of the individual tasks.

\begin{figure}[htp]
	\centering \subfigure[]{\includegraphics[height=1in]{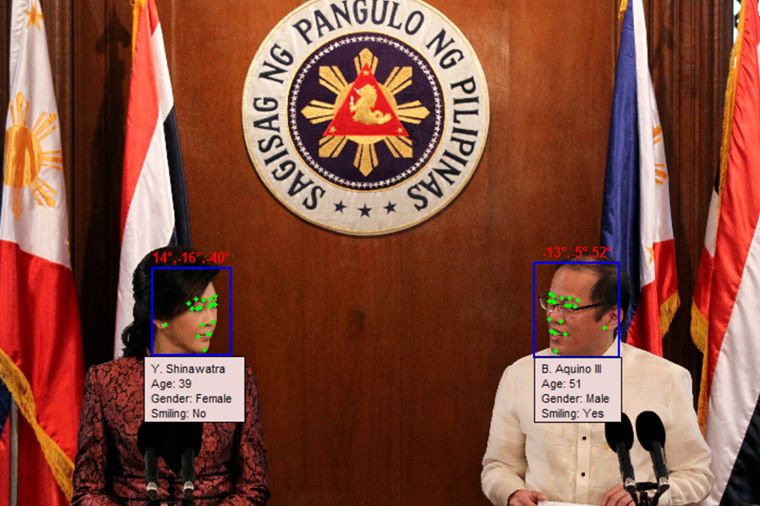}}~
        \subfigure[]{\includegraphics[height=1in]{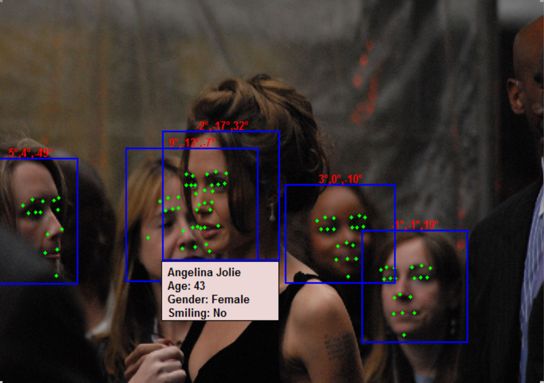}}  
        
    \caption{Sample outputs from the All-In-One
		Face~\cite{ranjan2016all} CNN for IJB-A ~\cite{IJBA}.} 
        \label{exp:ultraface_sample}
\end{figure}

Due to the lack of a single dataset which contains annotations for each task, various sub-networks
were trained with different datasets. These sub-networks share parameters among them. This ensures
that the shared parameters adapt to all the tasks instead of being task-specific. These sub-networks
are fused into a single All-in-One Face CNN at test time. Table~\ref{method:training_data} gives some
details of the datasets used for training All-in-One Face CNN. The complete network is trained end-to-end
using task-specific loss functions. Figure \ref{exp:ultraface_sample} shows some representative
outputs of the All-in-One Face CNN.

\begin{table}[htp!]
	\centering 
	\begin{tabular}{|c|c|c|}
		\hline
		Dataset                                     & Facial Analysis Task       & \# training samples \\ \hline\hline
		CASIA~\cite{yi_learning_2014}               & Identification, Gender     & 490,356 \\ \hline
		MORPH~\cite{1613043}                        & Age, Gender                & 55,608  \\ \hline
		IMDB+WIKI~\cite{rothe2015dex}               & Age, Gender                & 224,840 \\ \hline
		Adience~\cite{levi2015age}                  & Age                        & 19,370  \\ \hline
		CelebA~\cite{liu2015deep}                   & Smile, Gender              & 182,637 \\ \hline
		AFLW~\cite{tugraz:icg:lrs:koestinger11b}    & Detection, Pose, Fiducials & 20,342  \\ \hline
		Total                                       &                            & \textbf{993,153} \\ \hline
	\end{tabular}
    \vspace{2mm}
\caption{Datasets used for training All-In-One Face.}
	\label{method:training_data}   
\end{table}

The All-In-One network architecture uses the pre-trained face identification network from
Sankaranarayanan~\emph{et al.}~\cite{swami_btas_2016}, which contains seven convolutional layers
followed by three fully connected layers. This network is used as a backbone to train the face
identification task. The parameters from the first six convolutional layers of this network are
shared among the other face-related tasks as shown in figure \ref{fig:system_overview}. A CNN
pre-trained on face identification task provides better initialization for a generic face analysis
task, since the filters retain discriminative face information.

The face-related tasks used for training are divided into two categories: 1) subject-independent
tasks of face detection, landmarks localization, pose estimation, and smile prediction; 2)
subject-dependent tasks of gender classification, age estimation and face recognition. The
subject-dependent tasks require fine-grained features for classification, and hence need
semantically stronger features present in the deeper layers of the network. On the other hand,
subject-independent tasks are more localization oriented which need the features from shallower
layers of the network. Keeping these requirements in mind, the subject-independent features are
pooled from the first, third and fifth convolutional layers, while subject-dependent features are
pooled from deeper layers (see Fig.~\ref{fig:system_overview}).

Although All-In-One Face~\cite{ranjan2016all} provides outputs for seven different face-related
tasks, we use only the facial keypoints generated by this network in our face recognition pipeline.
Once we obtain the keypoints for every face in an image or a video frame, we align the faces to
normalized the canonical coordinates to mitigate the effects of in-plane rotation and scaling. These
aligned faces are then passed to the face recognition module for further processing.

\subsection{Face Identification and Verification}
\label{sec:face_id_ver}
In this subsection, we discuss our approach for face identification and verification. We use Crystal
Loss~\cite{ranjan2018crystal} to train deep networks for the task of face classification. Identity
features are then extracted for a face image from the penultimate layer of the trained networks.
These features undergo triplet embedding~\cite{swami_btas_2016} and fusion to generate a template
representation for an identity.

\subsubsection{Crystal Loss for Training CNNs}
Until recently, almost all face identification/verification networks were trained in the classification
setting using a softmax loss function. However, softmax loss is not ideal for training networks for
face representation. Softmax loss does not optimize the features to be similar for faces of the same
person and dissimilar for faces of different people. This leads to reduced performance. To alleviate
these issues Ranjan \emph{et al.} introduced the crystal loss function \cite{ranjan2018crystal} for training
networks used for unconstrained face verification and identification. The main idea behind this is to
constrain the features to lie on a hypersphere of a fixed radius. This ensures that the features
learnt are well-separated for different identities but close for same identities. Scaling the
features means that every image has a feature with the same scale. Contrast this with the softmax
loss
where high quality images usually give a feature with higher norm. This causes softmax loss to give more
attention to good quality images. This issue is also alleviated by the crystal loss which gives
equal importance to high and low quality images. 

The objective function for a network trained with crystal loss can be written as:
\begin{equation}
    \begin{aligned}
        & \text{minimize} & & -\frac{1}{M}\sum_{i=1}^M \log \frac{e^{W^T_{y_i}f(\mathbf{X}_i) +
b_{y_i}}}{\sum_{j=1}^C e^{W^T_j f(\mathbf{X}_i) + b_j}}\\
& \text{subject to} & & \lVert f(\mathbf{X}_i) \rVert_2 = \alpha, \forall i = 1,2,\dots M,
    \end{aligned}
\end{equation}
where $\mathbf{X}_i$ is the input image with label $y_i$, $M$ is the batch size, $f(\mathbf{X}_i)$
is the feature representation from the network, $C$ is the number of classes, $W$ and $b$ are
the weights of the classification layer of the network, and $\alpha$ is the scale of the feature
representation. 

This objective can be easily integrated into the network by simply normalizing the feature and
scaling it by $\alpha$ and applying softmax loss over this scaled representation. This module is
fully differentiable and can be inserted into any network trained using softmax loss.

\subsubsection{Training Datasets}
We use the Universe face dataset from \cite{bansal2018deep} for training our face representation
networks. This is a combination of UMDFaces
images \cite{UMDFaces}, UMDFaces video frames \cite{bansal2017s}, and curated MS-Celeb-1M
\cite{MS1M}. The Universe dataset contains about 5.6 million images of about 58,000 identities. This
includes about 3.5 million images from MS-Celeb-1M, 1.8 million video frames from UMDFaces videos,
and 300,000 images from UMDFaces. This dataset has the advantage of being a combination of
different datasets which makes networks trained using this dataset generalize better.
Another advantage is that it contains both still images and video frames. This makes the networks
more robust for test datasets which contain both.

\subsubsection{Face Representation}
\label{sec:face_rep}
We use two networks for feature representation. We do a score-level fusion of the scores obtained
from these networks. Using an ensemble of networks leads to more robust representations and
better performance. We next describe the two networks along with their respective training
details. These two networks are based on a ResNet-101
\cite{he2016deep}, and Inception ResNet-v2 \cite{inceptionresnet}.

For pre-processing the detected faces, we crop and resize the aligned faces to each network's input
dimensions. For data augmentation, we apply random horizontal flips to the input images.
\newline
\newline
\textbf{ResNet-101 (RG1)}
\newline
        We train a ResNet-101 deep convolutional neural network with PReLU activations after every
        convolution layer. Since we use the Universe dataset for training the network, we use a
        $58,000$-way classification layer with crystal loss. For this network, we set the $\alpha$
        parameter to $50$ and the batch size was $128$. The learning rate started at $0.1$ and was
        reduced by a factor of 0.2 after every $50k$ iterations. The network was trained for a total
        of $250k$ iterations. We use a $512$-D layer as the feature layer and use TPE
        \cite{swami_btas_2016} to find a 128-D embedding with was trained with the UMDFaces dataset.
\newline
\newline
\textbf{Inception ResNet-v2 (A)}
\newline
        The Inception ResNet-v2 network was also trained with the Universe dataset. This network has
        244 convolution layers. We add a $512$-D feature layer after these and then a final
        classification layer. We again use crystal loss with $\alpha = 40$. The initial learning
        rate was $0.1$ and reduced by a factor of $0.2$ after every $50k$ iterations. We trained the
        network for $120k$ iterations with a batch-size of $120$ on 8 NVIDIA Quadro P6000 GPUs. We
        resize the inputs to $299\times299$. Similar to the ResNet, we use UMDFaces to train a final
        128-D embedding with TPE.

\subsubsection{Feature Fusion}
\label{sec:fusion}
\textbf{Template Feature}
\newline
For both face verification and identification, we need to compare template features. To obtain
feature vectors for a template, we first average all the features for a media in the template. We
further average these media-averaged features to get the final template feature.
\newline
\newline
\textbf{Score-level Fusion}
\newline
To get the similarity between two templates, we average the similarities obtained by our two
networks.

%% file: dpssd.tex
\subsection{Deep Pyramid Single Shot Face Detector} 
\label{sec:dpssd}

\begin{figure*}[htp!]
	\begin{center}	
		\includegraphics[width=1\textwidth]{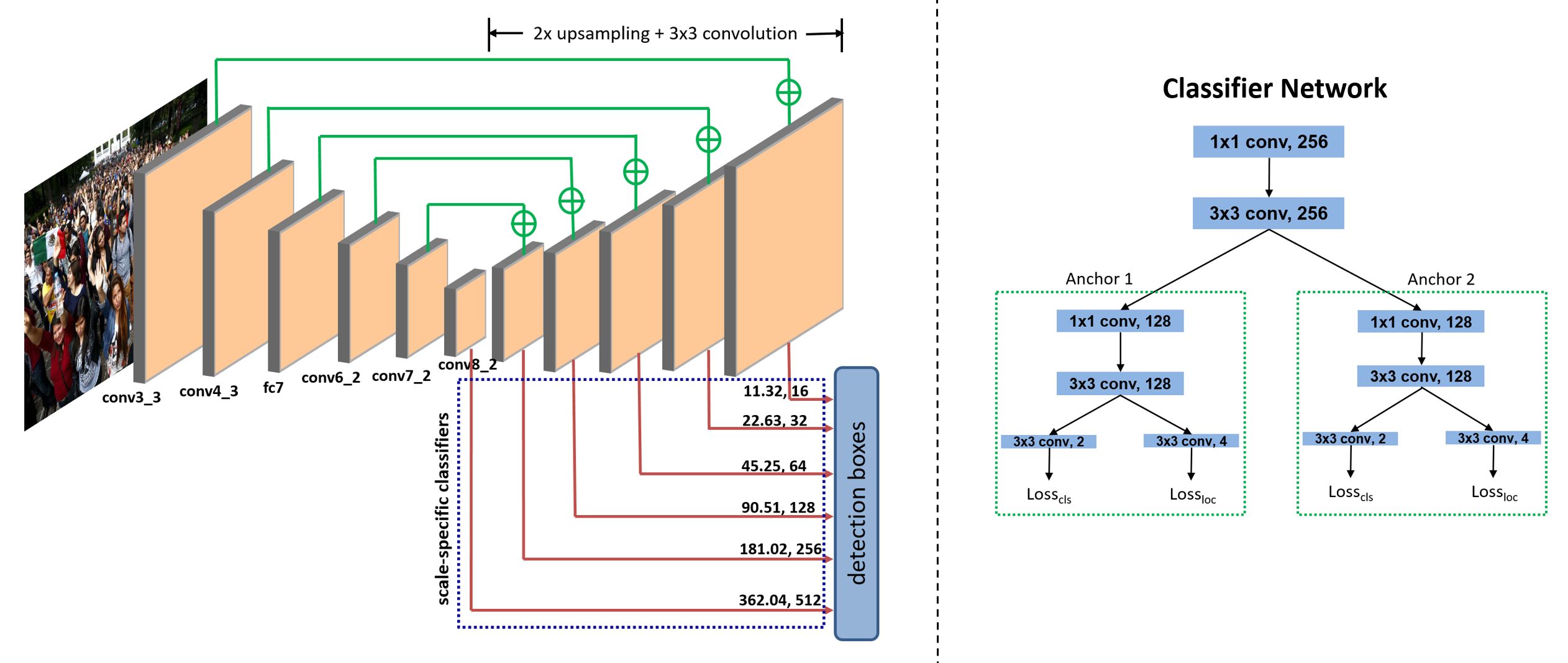}
	\end{center}
	\caption{The network architecture for the proposed Deep Pyramid Single Shot Face Detector
        (DPSSD). Starting from the base SSD~\cite{liu2016ssd} network, we add upsampling layers in
        an hourglass manner \cite{hourglass} to generate rich contextual features for face detection. The faces are
        pooled from six different layers of the network, with 2 scales at each layer. The numbers on
        the red arrows denote the anchor sizes for a given layer. The classifier network generates
        the face detection probability scores as well as the normalized bounding box offsets for
    every anchor (shown on right).}
	\label{fig:dpssd_arch}
	\vspace*{-4mm}
\end{figure*}

We propose a novel DCNN-based face detector, called Deep Pyramid Single Shot Face Detector (DPSSD),
that is fast and capable of detecting faces at a large variety of scales. It is especially good at
detecting tiny faces. Since face detection is a special case
of generic object detection, many researchers have used an off-the-shelf object detector and
fine-tuned it for the task of face detection~\cite{jiang2016face}. However, in order to design an
efficient face detector, it is crucial to address the following differences between the tasks of
face and object detection. First, the faces can occur at a much lower scale/size in an
image compared to a general object. 
Typically, object detectors are not designed to detect at such a low resolution which is required for the
task of face detection. Second, variations in the aspect ratio of faces are much less compared
to those in a typical object. As faces incur less structural deformations compared to objects, they
do not need any additional processing incorporated in object detection algorithms to handle
multiple aspect ratios. We design our face detector to addresses these points.


We start with the Single Shot Detector (SSD)~\cite{liu2016ssd} trained on the truncated
VGG-16~\cite{simonyan2014very} network for the task of object detection. SSD~\cite{liu2016ssd} has a
speed advantage over other object detectors like Faster R-CNN~\cite{ren2015faster} since it is
single stage and does not use proposals. The SSD approach is fully convolutional and generates a
fixed number of bounding boxes and scores for the presence of faces. Additional convolutional layers
are added to the end of the truncated VGG-16~\cite{simonyan2014very} to detect objects at multiple
scales. The objects are detected from multiple feature layers using different convolutional models
for each layer. We modify the SSD~\cite{liu2016ssd} architecture and the approach in such a way that
it is able to detect tiny faces efficiently. Fig.~\ref{fig:dpssd_arch} shows the overall
architecture of the proposed DPSSD face detector.
\\

\noindent
\textbf{Anchor pyramid with fixed aspect-ratio:} In order to detect faces at multiple scales, we
leverage the feature pyramid structure inbuilt in the DCNN. We resize the input image such that the
side with minimum length has a dimension of $512$. After every convolutional block, max pooling is
performed which reduces the dimension of feature maps by half and doubles the stride. For instance,
the feature maps at conv$3\_3$ layer have a minimum spatial dimension of $128$. Additionally, a unit
stride in this layer corresponds to $4$ pixels stride in the original image. As shown in table
\ref{tbl:dpssd}, initial layers of a
DCNN have low stride in feature maps, which is beneficial for detecting tiny faces since small size
anchors can be matched with high Jaccard overlap of $0.5$. However, features from these layers
have less discriminative ability. On the other hand, features from deeper layers are semantically
stronger, but do not provide good  spatial localization because of the large stride value. Hence, we
choose the anchor sizes and the corresponding feature maps for generating detections with an optimal
combination of low spatial stride and highly discriminative features.

\begin{table}[H]
	\begin{center}
		\begin{tabular}{|c|c|c|c|}				
			\hline
			Layer & Stride (pixels) & Anchor-Sizes (pixels) & \#boxes \\
			\hline
			conv$3\_3$ & 4 & 16/$\sqrt{2}$, 16 & 32768\\
			\hline
			conv$4\_3$ & 8 & 32/$\sqrt{2}$, 32 & 8192\\
			\hline
			fc$7$ & 16 & 64/$\sqrt{2}$, 64 & 2048\\
			\hline
			conv$6\_2$ & 32 & 128/$\sqrt{2}$, 128 & 512\\
			\hline
			conv$7\_2$ & 64 & 256/$\sqrt{2}$, 256 & 128\\
			\hline
			conv$8\_2$ & 128 & 512/$\sqrt{2}$, 512 & 32\\
			\hline
		\end{tabular}
		\vspace{2mm}
		\caption{Statistics for different layers of DPSSD. The sizes of the two anchors and the stride are measured in pixels.}
		\label{tbl:dpssd}
		
	\end{center}
\end{table}
\vspace{-5mm}

We choose $12$ anchor boxes, each at a different scale. The largest anchor box has a size of $512$.
Each anchor box maintains a scale factor of $\sqrt{2}$ with its next lower level in the hierarchy.
We apply these anchor boxes to generate detections from $6$ different feature maps (see
Table~\ref{tbl:dpssd}). Small-sized anchor boxes are applied to shallower feature maps while
large-sized anchor boxes are applied to deeper feature maps. Unlike SSD~\cite{liu2016ssd}, we make
use of the conv$3\_3$ layer for generating the detections since it has a high spatial resolution of
$128$. This helps us in detecting tiny faces of size as low as $8$ pixels.

We fix the aspect ratio of every anchor box to the mean aspect-ratio for face ($0.8$). We compute
this value from the WIDER Face~\cite{yang2016wider} training dataset. For a given anchor size $a$,
the anchor box $m \times n$ is calculated as:

\begin{equation}
m = \frac{a}{\sqrt{0.8}}, ~~~~~n = a \times \sqrt{0.8},
\end{equation}
where $m$ is the width and $n$ is the height of the anchor box. Detection scores and bounding box
offsets are provided at each location of the feature map for a given anchor box. Feature maps with
larger spatial resolution result in more detection boxes. The number of detection boxes
generated by every anchor layer for an image of size $512 \times 512$, is also provided in
table~\ref{tbl:dpssd}. The conv$3\_3$ layer outputs the largest number of boxes since it has a
spatial resolution of $128 \times 128$. All of these generated boxes are passed through the classifier
at the time of training.
\\

\noindent
\textbf{Contextual Features from upsampling layers:} It has been established that contextual
information is useful for detecting tiny faces~\cite{hu2016finding}. Although features from the
conv$3\_3$ layer have appropriate spatial resolution for tiny face detection, they are neither
semantically strong nor they contain contextual information. In order to provide contextual
information, we add a stack of bilinear upsampling and convolution layer at the end of the
SSD~\cite{liu2016ssd} network. The $6$ chosen layers (Table~\ref{tbl:dpssd}) are then added
element-wise to these upsampled layers (see Fig.~\ref{fig:dpssd_arch}). Thus, the features
become rich in both semantics and localization. The final detection boxes are generated from these
upsampled layers using the anchor box matching technique.

Every output level generates two sets of detections, one for each anchor box corresponding to the
given layer. A classifier network (see Fig.~\ref{fig:dpssd_arch} right) is attached to all the $6$
output feature maps, that provides the classification probabilities and bounding box offsets
corresponding to each of the $12$ anchor boxes. The classifier network is branched into two to
handle each anchor box separately. These branches are further subdivided into classification and
regression subnetworks.

\subsubsection{Training}
We use the training set of WIDER Face~\cite{yang2016wider} dataset to train our face detector. The
network is initialized with the SSD~\cite{liu2016ssd} model for object detection. The new layers
that are added are initialized randomly. We use a batch size of $8$. The initial learning rate is
set to $0.001$ which is decreased by $0.5$ after $30k$, $50k$ and $60k$ iterations. Training is
carried out till $70k$ iterations.  The matching strategy is similar to SSD~\cite{liu2016ssd}. A
location in the predictor feature map is labeled as positive class ($y_{c} = 1$) if the anchor box
for that location has an Intersection-over-Union (IoU) overlap of $0.5$ or more with any ground
truth face. All the other locations are labeled as negative class ($y_{c} = 0$). For all the
positive classes, we also perform bounding box regression. We use the binary cross-entropy loss for
face classification and smooth-L1 loss for bounding box regression. The overall loss ($L$) is a
weighted sum of classification loss ($L_{cls}$) and regression loss ($L_{loc}$) as shown
in~(\ref{eq:L_cls}), (\ref{eq:L_loc}) and (\ref{eq:L_all}). We use Caffe~\cite{jia_caffe_2014}
library to train our network.

\begin{equation}
\label{eq:L_cls}
L_{cls} =  -y_{c} \cdot \log(p_{c}) - (1 - y_{c}) \cdot \log(1- p_{c}),
\end{equation}

\begin{equation}
\label{eq:L_loc}
L_{loc} = \sum \limits_{i \in \{x,y,w,h\}}~\mathrm{smooth}_{L1}~(t_{i} - v_{i}),
\end{equation}

\begin{equation}
\label{eq:L_all}
L =  L_{cls} + \lambda \cdot y_{c} \cdot L_{loc},
\end{equation}

where $y_{c}$ is the class label, $p_{c}$ is the softmax probability obtained from the network, $v =
\{v_{x}, v_{y}, v_{w}, v_{h}\}$ denote the ground-truth normalized bounding box regression targets
while $t = \{t_{x}, t_{y}, t_{w}, t_{h}\}$ are the bounding box offsets predicted by the network.
The value of $\lambda$ is chosen to be $1$. The $\mathrm{smooth}_{L1}$ loss is defined
in~(\ref{eq:smooth_l1}).

\begin{equation}
\label{eq:smooth_l1}
\mathrm{smooth}_{L1}(x) = \begin{cases}
0.5x^{2} &\quad\text{if } |x|<1 \\
|x| - 0.5 &\quad\text{otherwise} \\  
\end{cases}
\end{equation}

The total number of detection boxes generated from an image is $43,680$. Out of these, only a few
boxes (around $10$-$50$) correspond to the positive class while others form the negative class. To
avoid this large class imbalance we select only a few negative boxes such that the ratio of positive
to negative class is $1:3$. We use hard negative mining to select these negative boxes as proposed
in~\cite{liu2016ssd}. We use the data augmentation technique proposed in~\cite{liu2016ssd} to make
the detector more robust to various face sizes.

\subsubsection{Testing}
\begin{figure}[h]
	\centering
	\includegraphics[width=0.5\textwidth]{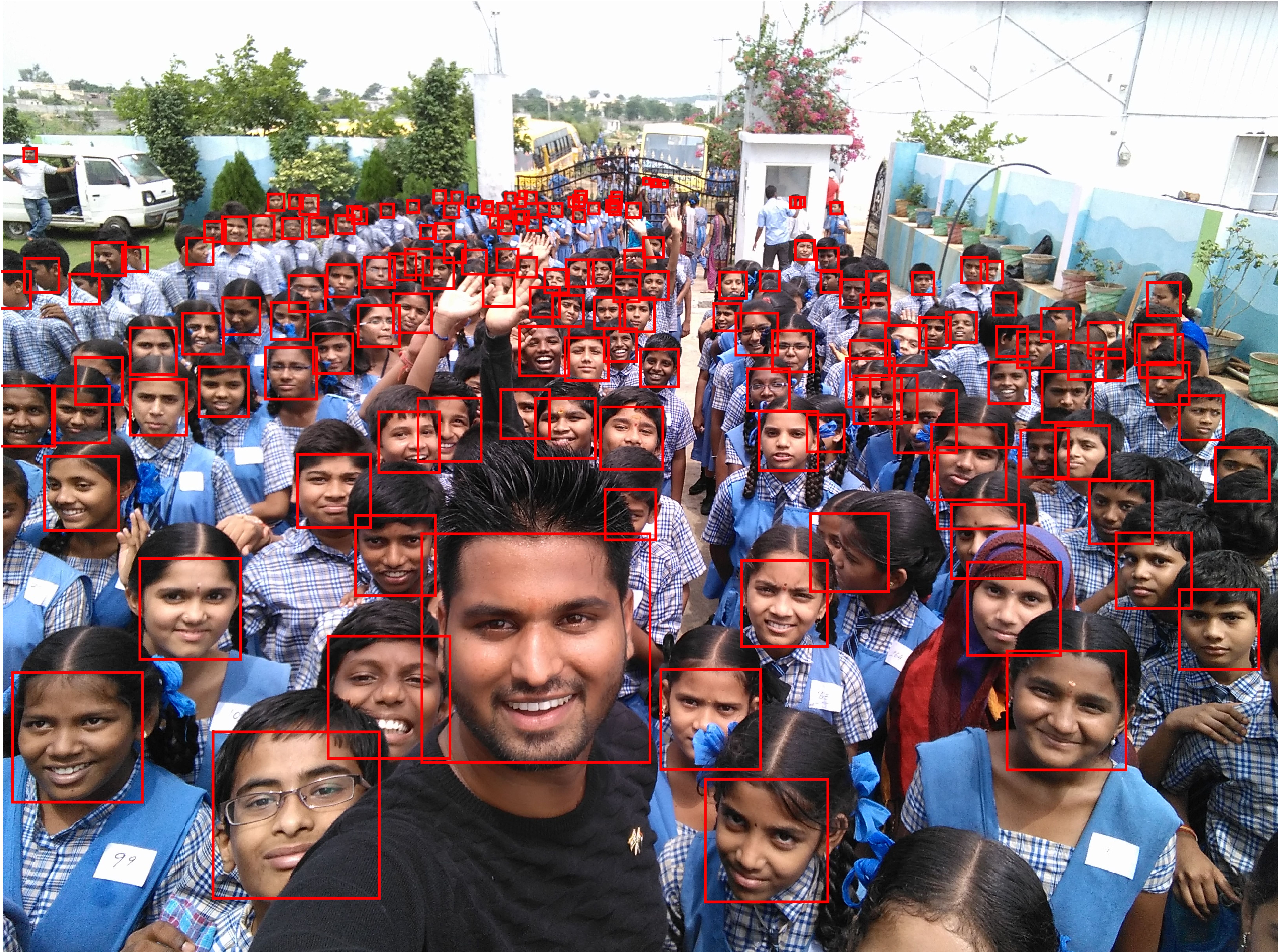}
	\caption{A sample output for our proposed DPSSD.}
	\label{fig:sample_dpssd}	
\end{figure}

At test time, the input image is resized such that the minimum side has the dimension of $512$
pixels. The aspect ratio of the image is not changed. The image is then passed through the trained
DPSSD face detector to get the detection scores and bounding box co-ordinates for different
locations in the image. Non-maximum suppression (NMS) with threshold of $0.6$ is used to filter out
the redundant boxes. Since the outputs are generated in a single pass of the network, the total
processing time is very low ($100ms$). To further improve the detection performance, we construct
the image pyramid as discussed in HR~\cite{hu2016finding} face detector. A sample face detection
output using the proposed DPSSD is shown in Fig.~\ref{fig:sample_dpssd}. Performance evaluations of
different face detection datasets are discuss in Section~\ref{sec:experiments}.

For the proposed face recognition pipeline, we use the results of both DPSSD and SSD for faces, as
our detectors to capture faces across as many scales as possible.

%% file: experiments.tex
\section{Experimental Results}
\label{sec:experiments}
In this section, we first report face detection results for the proposed detector on four datasets.
We also report experimental results for face identification and verification on four challenging
evaluation datasets, \emph{viz.}, IJB-A \cite{IJBA}, IJB-B \cite{IJBB}, IJB-C \cite{IJBC}, and the
IARPA Janus Challenge Set 5 (CS5). We show that the proposed system achieves state-of-the-art or near results
on most of the protocols. In the following sections we describe the evaluation datasets and
protocols. We also describe the changes to the system we made if there are any. 

\input{dpssd_exp.tex}

Table \ref{tab:all_results} lists the locations of the results for different datasets and
verification and identification evaluation tasks.

\begin{table}
    \centering
    \resizebox{0.5\textwidth}{!}
    {
    \begin{tabular}{|c|c|c|c|c|}
        \hline
        \textbf{Task} & \textbf{IJB-A} & \textbf{IJB-B} & \textbf{IJB-C} & \textbf{CS5} \\
        \hline
        1:1 Verification & Table \ref{tab:ijba_verification} & Table \ref{tab:ijbb_verification} &
        Table \ref{tab:ijbc_verification} & Table \ref{tab:cs5_verification}\\
        \hline
        1:N Search & Table \ref{tab:ijba_identification} & Table \ref{tab:ijbb_identification} &
        Table \ref{tab:ijbc_identification} & Table \ref{tab:cs5_identification}\\
        \hline
        Wild Probe Search & - & - & Table \ref{tab:ijbc_task9} & Table \ref{tab:cs5_task9}\\
        \hline
    \end{tabular}
    }
    \vspace{2mm}
    \caption{Locations of all results for face verification and identification.}
    \label{tab:all_results}
\end{table}

In the following sections, ROC curves are used to measure the performance of verification (1:1
matching) methods and CMC scores are used for evaluating identification (1:N search). The IJB-A,
IJB-B, IJB-C, and CS5 datasets contain a gallery and a probe which leads to evaluation using all
positive and negative pairs. This is different from LFW \cite{huang_labeled_2008} and YTF
\cite{wolf2011face} where only a few negative pairs are used to evaluate verification performance.
Another difference between LFW/YTF and the evaluation datasets here is the inclusion of templates
instead of only single images. A template is a collection of images and video frames of a subject.
These datasets are much more challenging than older datasets due to extreme variations in pose,
illumination, and expression.

\subsection{IJB-A}
The IJB-A dataset contains 500 subjects with 5,397 images and 2,042 videos split into 20,412 frames.
This dataset is a very difficult dataset owing
to the presence of extreme pose, viewpoint, resolution, and illumination variations. Additionally,
mixing still images and video frames causes difficulties for models trained with only one of these
modalities due to domain shift. An identity in this dataset is represented as a template. A template
is a set of face images instead of just a single image. This set can contain still images and video
frames. Also note that each subject can have multiple templates in the dataset. The evaluation for
this dataset contains for 1:1 verification and 1:N mixed search protocols. Table
\ref{tab:ijba_b_task_desc} gives brief descriptions of the two tasks. The dataset is divided
into 10 splits, each with $333$ randomly selected subjects for training and $167$ subjects for
testing. To represent each template, we generate a common representation by fusing features of all
the faces in the template. We compute the similarity scores using all the networks and then do a
score-level fusion as described in \ref{sec:fusion}. Table \ref{tab:ijba_verification} gives the
results from our system for the verification task for IJB-A and table \ref{tab:ijba_identification}
gives the results for 1:N mixed search. We achieve state-of-the-art results for every
setting. 

\begin{table*}
    \centering
        \begin{tabular}{|c|c|}
            \hline
            \textbf{Task} & \textbf{Desciption}\\
            \hline
            1:1 Verification & Verify if the given pair of templates belong to the same subject.
            Templates are comprised of mixed media (frames and stills)\\
            \hline
            1:N Mixed Search & Open set identification protocol using mixed media (frames and stills)
            as probe and two galleries G1, and G2.\\
            \hline
        \end{tabular}
    \vspace{2mm}
    \caption{IJB-A and IJB-B task descriptions}
    \label{tab:ijba_b_task_desc}
\end{table*}

\begin{table}
    \centering
        \begin{tabular}{|c|c|c|c|c|}
            \hline
            & \multicolumn{4}{|c|}{True Accept Rate (\%) @ False Accept Rate}\\
            \hline
            \textbf{Method} & 0.0001 & 0.001 & 0.01 & 0.1 \\
            \hline
            Casia \cite{wang_face_2015} & - & 51.4 & 73.2 & 89.5 \\
            \hline
            Pose \cite{AbdAlmageed_pose_2016} & - & - & 78.7 & 91.1 \\
            \hline
            NAN \cite{yang_neural_2016} & - & 88.1 & 94.1 & 97.8 \\
            \hline
            3d \cite{masi_neural_2016} & - & 72.5 & 88.6 & - \\
            \hline
            DCNN$_{fusion}$ \cite{chen2016end} & - & 76.0 & 88.9 & 96.8 \\
            \hline
            DCNN$_{tpe}$ \cite{swami_btas_2016} & - & 81.3 & 90.0 & 96.4 \\
            \hline
            DCNN$_{all}$ \cite{ranjan2016all} & - & 78.7 & 89.3 & 96.8 \\
            \hline
            All + TPE \cite{ranjan2016all} & - & 82.3 & 92.2 & 97.6 \\
            \hline
            TP \cite{crosswhite_template_2016} & - & - & 93.9 & - \\
            \hline
            RX101$_{l2+tpe}$ \cite{ranjan2017l2} & 90.9 & 94.3 & 97.0 & 98.4 \\
            \hline
            \hline
            Ours$_A$ & 91.7 & 95.3 & 96.8 & 98.3 \\
            \hline
            Ours$_{RG1}$ & 91.4 & 94.8 & \textbf{97.1} & \textbf{98.5} \\
            \hline
            Fusion (Ours) & \textbf{92.1} & \textbf{95.2} & 96.9 & 98.4 \\
            \hline
        \end{tabular}
    \vspace{2mm}
    \caption{IJB-A Verification. A is our Inception ResNet-v2 model and RG1 is our ResNet-101 model.}
    \label{tab:ijba_verification}
\end{table}

\begin{table*}
    \centering
        \begin{tabular}{|c|c|c|c|c|c|}
            \hline
            & \multicolumn{2}{|c|}{TPIR (\%) @ FPIR} & \multicolumn{3}{|c|}{Retrieval Rate (\%)}\\
            \hline
            \textbf{Method} & 0.01 & 0.1 & Rank=1 & Rank=5 & Rank=10 \\
            \hline
            Casia \cite{wang_face_2015} & 38.3 & 61.3 & 82.0 & 92.9 & - \\
            \hline
            Pose \cite{AbdAlmageed_pose_2016} & 52.0 & 75.0 & 84.6 & 92.7 & 94.7 \\
            \hline
            BL \cite{aruni_pose_2016} & - & - & 89.5 & 96.3 & - \\
            \hline
            NAN \cite{yang_neural_2016} & 81.7 & 91.7 & 95.8 & 98 & 98.6 \\
            \hline
            3d \cite{masi_neural_2016} & - & - & 90.6 & 96.2 & 97.7 \\
            \hline
            DCNN$_{fusion}$ \cite{chen2016end} & 65.4 & 83.6 & 94.2 & 98.0 & 98.8 \\
            \hline
            DCNN$_{tpe}$ \cite{swami_btas_2016} & 75.3 & 83.6 & 93.2 & - & 97.7 \\
            \hline
            DCNN$_{all}$ \cite{ranjan2016all} & 70.4 & 83.6 & 94.1 & - & 98.8 \\
            \hline
            ALL + TPE \cite{ranjan2016all} & 79.2 & 88.7 & 94.7 & - & 98.8 \\
            \hline
            TP \cite{crosswhite_template_2016} & 77.4 & 88.2 & 92.8 & - & 98.6 \\
            \hline
            RX101$_{l2+tpe}$ \cite{ranjan2017l2} & 91.5 & 95.6 & 97.3 & - & 98.8 \\
            \hline
            \hline
            Ours$_A$ & 91.4 & 96.1 & 97.3 & 98.2 & 98.5 \\
            \hline
            Ours$_{RG1}$ & 91.6 & 96.0 & 97.4 & 98.5 & 98.9 \\
            \hline
            Fusion (Ours) & \textbf{92.0} & \textbf{96.2} & \textbf{97.5} & \textbf{98.6} & \textbf{98.9} \\
            \hline
        \end{tabular}
    \vspace{2mm}
    \caption{IJB-A 1:N Mixed Search. A is our Inception ResNet-v2 model and RG1 is our ResNet-101 model.}
    \label{tab:ijba_identification}
\end{table*}

\subsection{IJB-B}
The IJB-B dataset \cite{IJBB}, which extends IJB-A, contains about $22,000$ still images and
$55,000$ video frames spread over $1,845$ subjects. Evaluation is done for the same tasks as IJB-A,
\emph{viz.}, 1:1 verification, and 1:N identification (table \ref{tab:ijba_b_task_desc}). The IJB-B
verification protocol consists of $8,010,270$ pairs between
templates in the galleries (G1 and G2) and the probe templates. Out of these 8 million are
impostor pairs and the rest $10,270$ are genuine comparisons. Table \ref{tab:ijbb_verification} gives
verifications results and table \ref{tab:ijbb_identification} gives identification results. 

\begin{table*}
    \centering
        \begin{tabular}{|c|c|c|c|c|c|c|c|}
            \hline
            & \multicolumn{7}{|c|}{True Accept Rate (\%) @ False Accept Rate}\\
            \hline
            \textbf{Method} & $10^{-7}$ & $10^{-6}$ & $10^{-5}$ & $10^{-4}$ & $10^{-3}$ & $10^{-2}$ & $10^{-1}$ \\
            \hline
            GOTS \cite{IJBB} & - & - & - & 16.0 & 33.0 & 60.0 & - \\
            \hline
            VGGFaces \cite{wen2016discriminative} & - & - & - & 55.0 & 72.0 & 86.0 & - \\
            \hline
            FPN \cite{chang2017faceposenet} & - & - & - & 83.2 & 91.6 & 96.5 & - \\
            \hline
            Light CNN-29 \cite{wu2018light} & - & - & - & 87.7 & 92.0 & 95.3 & - \\
            \hline
            VGGFace2 \cite{vggface2} & - & - & 70.5 & 83.1 & 90.8 & 95.6 & - \\
            \hline
            Center Loss Features \cite{wen2016discriminative} & \textbf{8.8} & 31.0 & 63.6 & 80.7 & 90.0 & 95.1 & 98.4 \\
            \hline
            \hline
            Ours$_A$ & 2.9 & 27.7 & 61.6 & 89.1 & 94.3 & 97.0 & 98.7\\
            \hline
            Ours$_{RG1}$ & 6.2 & \textbf{48.4} & \textbf{80.4} & 89.8 & 94.4 & 97.2 & 98.9\\
            \hline
            Fusion (Ours) & 4.4 & 45.6 & 77.8 & \textbf{90.3} & \textbf{94.6} & \textbf{97.3} & \textbf{98.9}\\
            \hline
        \end{tabular}
    \vspace{2mm}
    \caption{IJB-B Verification. A, RG1, and Fusion are our Inception ResNet-v2,
    ResNet-101, and Fused features respectively.}
    \label{tab:ijbb_verification}
\end{table*}

\begin{table*}
    \centering
        \begin{tabular}{|c|c|c|c|c|c|}
            \hline
            & \multicolumn{2}{|c|}{TPIR (\%) @ FPIR (For G1, G2)} & \multicolumn{3}{|c|}{Retrieval
            Rate (\%) (For G1, G2)}\\
            \hline
            \textbf{Method} & 0.01 & 0.1 & Rank=1 & Rank=5 & Rank=10 \\
            GOTS \cite{IJBB} & - & - & 42.0 & - & 62.0 \\
            \hline
            VGGFace \cite{wen2016discriminative} & - & - & 78.0 & - & 89.0 \\
            \hline
            FPN \cite{chang2017faceposenet} & - & - & 91.1 & - & 96.5 \\
            \hline
            Light CNN-29 \cite{wu2018light} & - & - & 91.9 & 94.8 & - \\
            \hline
            VGGFace2 \cite{vggface2} & 74.3 & 86.3 & 90.2 & 94.6 & 95.9 \\
            \hline
            Center Loss Features \cite{wen2016discriminative} & 75.5, 67.7 & 87.5, 82.8 & 92.2, 86.0 & 95.4, 92.5 & 96.2,
            94.4 \\
            \hline
            \hline
            Ours$_A$ & 83.1, 75.5 & 93.6, 89.3 & 95.5, 90.8 & 97.5, 94.2 &
            98.0, 95.8  \\
            \hline
            Ours$_{RG1}$ & 86.9, 78.6  & 94.0, 89.1  & 95.6, 91.5  & \textbf{97.7, 95.4}  & 98.0,
            \textbf{96.5}  \\
            \hline
            Fusion (Ours) & \textbf{88.2, 79.4} & \textbf{94.3, 89.7} & \textbf{95.8, 91.8} & 97.7,
            95.2 & \textbf{98.1}, 96.4  \\
            \hline
        \end{tabular}
    \vspace{2mm}
    \caption{IJB-B 1:N Mixed Search. A and RG1 are our Inception ResNet-v2 and ResNet-101 models respectively. Note that the retrieval rates for some past methods are average over G1 and G2.}
    \label{tab:ijbb_identification}
\end{table*}

\subsection{IJB-C}
The IJB-C evaluation dataset \cite{IJBC} further extends IJB-B. It contains $31,334$ still images
and $117,542$ video frames of $3,531$ subjects. In addition to the evaluations from IJB-B, this
dataset evaluates end-to-end recognition. Table \ref{tab:ijbc_task_desc} gives the descriptions of
the evaluated tasks for IJB-C. There are about $20,000$ genuine comparisons, and about
$15.6$ million impostor pairs in the verification protocol. For the 1:N mixed search protocol, there
are about $20,000$ probe templates. In table \ref{tab:ijbc_verification} we list the
results of our system for 1:1 verification. Similarly, in
table \ref{tab:ijbc_identification} we give results for 1:N mixed search. We also report the
1:N wild probe search results in table \ref{tab:ijbc_task9}.

\begin{table*}
    \centering
        \begin{tabular}{|c|c|}
            \hline
            \textbf{Task} & \textbf{Desciption}\\
            \hline
            1:1 Verification & Verify if the given pair of templates belong to the same subject.
            Templates are comprised of mixed media (frames and stills)\\
            \hline
            1:N Mixed Search & Open set identification protocol using mixed media (frames and stills)
            as probe and two galleries G1, and G2.\\
            \hline
            Wild Probe Search & Identify subjects of interest from a collection of still images and
            frames. This task also uses the two galleries G1, and G2.\\
            \hline
        \end{tabular}
    \vspace{2mm}
    \caption{IJB-C task descriptions}
    \label{tab:ijbc_task_desc}
\end{table*}

\begin{table*}
    \centering
        \begin{tabular}{|c|c|c|c|c|c|c|c|c|}
            \hline
            & \multicolumn{8}{|c|}{True Accept Rate (\%) @ False Accept Rate}\\
            \hline
            \textbf{Method} & $10^{-8}$ & $10^{-7}$ & $10^{-6}$ & $10^{-5}$ & $10^{-4}$ & $10^{-3}$ & $10^{-2}$ & $10^{-1}$ \\
            \hline
            Center Loss Features \cite{wen2016discriminative} & 36.0 & 37.6 & 66.1 & 78.1 & 85.3 & 91.2 & 95.3 & 98.2 \\
            \hline
            \hline
            Ours$_A$ & 16.5 & 19.5 & 43.6 & 77.6 & 91.9 & 95.6 & 97.8 &
            99.0\\
            \hline
            Ours$_{RG1}$ & \textbf{60.6} & \textbf{67.4} & \textbf{76.4} & 86.2 & 91.9 & 95.7 & 97.9 & 99.2\\
            \hline
            Fusion (Ours) & 54.1 & 55.9 & 69.5 & \textbf{86.9} & \textbf{92.5} & \textbf{95.9} &
            \textbf{97.9} & \textbf{99.2}\\
            \hline
        \end{tabular}
    \vspace{2mm}
    \caption{IJB-C Verification. A is our Inception ResNet-v2 model and RG1 is our ResNet-101 model. Fusion is the fusion of the two features.}
    \label{tab:ijbc_verification}
\end{table*}

\begin{table*}
    \centering
        \begin{tabular}{|c|c|c|c|c|c|}
            \hline
            & \multicolumn{2}{|c|}{TPIR (\%) @ FPIR (For G1, G2)} & \multicolumn{3}{|c|}{Retrieval
            Rate (\%) (For G1, G2)}\\
            \hline
            \textbf{Method} & 0.01 & 0.1 & Rank=1 & Rank=5 & Rank=10 \\
            \hline
            Center Loss Features \cite{wen2016discriminative} & 79.1, 75.3 & 86.4, 84.2 & 91.7, 89.8 & 94.6, 93.6 & 95.6,
            94.9\\
            \hline
            \hline
            Ours$_A$ & 87.7, 82.4 & 93.5, 91.0 & 95.7, 92.8 & 97.4, 95.4 &
            97.9, 96.4\\
            \hline
            Ours$_{RG1}$ & 88.0, 84.2 & 93.2, 90.6 & 95.9, 93.2 & 97.6, 96.1 & 98.1,
            \textbf{97.0}\\
            \hline
            Fusion (Ours) & \textbf{89.6, 85.0} & \textbf{93.8, 91.3} & \textbf{96.2, 93.6} &
            \textbf{97.7, 96.2} & \textbf{98.2}, 96.9\\
            \hline
        \end{tabular}
    \vspace{2mm}
    \caption{IJB-C 1:N Mixed Search. A and RG1 are our models described in section \ref{sec:face_rep}.}
    \label{tab:ijbc_identification}
\end{table*}

\begin{table*}
    \centering
        \begin{tabular}{|c|c|c|c|c|c|c|}
            \hline
            & \multicolumn{6}{|c|}{Retrieval Rate (\%) (For G1, G2)}\\
            \hline
            \textbf{Method} & Rank=1 & Rank=2 & Rank=5 & Rank=10 & Rank=20 & Rank=50 \\
            \hline
            Ours$_A$ & 91.1, 86.9 & 93.0, 89.0 & 94.8, 91.1 & 95.8, 92.5 &
            96.5, 93.8 & 97.4, 95.3\\
            \hline
            Ours$_{RG1}$ & 90.8, 86.3 & 93.0, 88.8 & 95.0, 91.1 & 96.0, 92.6 & 96.7, 93.9
            & 97.5, 95.5\\
            \hline
            Fusion (Ours) & \textbf{91.8, 87.5} & \textbf{93.6, 89.7} & \textbf{95.3, 91.6} &
            \textbf{96.3, 93.0} & \textbf{97.0, 94.4} & \textbf{97.7, 95.8}\\
            \hline
        \end{tabular}
    \vspace{2mm}
    \caption{IJB-C Wild Probe Search. Our models and fusion method are described in sections \ref{sec:face_rep} and \ref{sec:fusion}.}
    \label{tab:ijbc_task9}
\end{table*}

\subsection{CS5}
We evaluated on the (as-yet-unreleased to public) JANUS Challenge Set 5 dataset also. We give the task
descriptions for this dataset in table \ref{tab:cs5_task_desc}. This dataset consists of
$2,875,950$ still images. This dataset also provides a training set consisting of $235,616$ identity
clusters and $981,753$ images. Note that we did not use this training set for training our networks.
The still image verification protocol contains $547,131$ templates with $332,574$ genuine matches,
and $822,354,805$ imposter matches. For the 1:N identification task, there are $332,574$ probe
templates. Gallery, G1 has $1,106,778$ identity clusters and G2 has $1,107,779$ identity clusters.
Tables \ref{tab:cs5_verification}, \ref{tab:cs5_identification}, and \ref{tab:cs5_task9} give results for
1:1 verification, 1:N identification, and 1:N end-to-end identification respectively.

Note that we are unable to compare the performance of the proposed approach against other methods for IJB-C
and CS5 due to restrictions on publishing competitor's results.

\begin{table*}
    \centering
        \begin{tabular}{|c|c|}
            \hline
            \textbf{Task} & \textbf{Desciption}\\
            \hline
            1:1 Still Image Verification &  Templates are comprised of only still images.\\
            \hline
            1:N Still Image Identification & Open set identification protocol using still images
            as probe and two galleries G1, and G2 augmented with 1M distractors.\\
            \hline
            1:N end-to-end Still Image & Identify identity clusters of interest from a collection of
            still images. This task also uses the two galleries G1, and G2.\\
            \hline
        \end{tabular}
    \vspace{2mm}
        \caption{IARPA Janus Challenge Set 5 (CS5) task descriptions}
    \label{tab:cs5_task_desc}
\end{table*}

\begin{table*}
    \centering
        \begin{tabular}{|c|c|c|c|c|c|c|c|c|}
            \hline
            & \multicolumn{8}{|c|}{True Accept Rate (\%) @ False Accept Rate}\\
            \hline
            \textbf{Method} & $10^{-8}$ & $10^{-7}$ & $10^{-6}$ & $10^{-5}$ & $10^{-4}$ & $10^{-3}$ & $10^{-2}$ & $10^{-1}$ \\
            \hline
            Ours$_A$ & 52.44 & 78.99 & 94.88 & 97.34 & 98.18 & 98.74 & 99.28 & 99.75 \\
            \hline
            Ours$_{RG1}$ & 71.52 & 89.68 & 95.20 & 97.28 & 98.19 & 98.79 & 99.36 & 99.78 \\
            \hline
            Fusion (Ours) & 70.72 & 90.74 & 95.80 & 97.49 & 98.25 & 98.80 & 99.35 & 99.78\\
            \hline
        \end{tabular}
    \vspace{2mm}
    \caption{CS5 1:1 Verification. A and RG1 are our Inception ResNet-v2 and ResNet-101 models respectively. Both of these models are trained with Crystal Loss.}
    \label{tab:cs5_verification}
\end{table*}

\begin{table*}
    \centering
        \begin{tabular}{|c|c|c|c|c|c|c|c|c|}
            \hline
            & \multicolumn{4}{|c|}{TPIR (\%) @ FPIR} & \multicolumn{4}{|c|}{Retrieval Rate (\%)}\\
            & \multicolumn{4}{|c|}{(Average for G1 and G2)} & \multicolumn{4}{|c|}{(Average for G1 and G2)}\\
            \hline
            \textbf{Method} & 0.0001 & 0.001 & 0.01 & 0.1 & Rank=1 & Rank=5 & Rank=10 & Rank=20 \\
            \hline
            Fusion (Ours) & 29.96 & 75.90 & 86.76 & 95.59 & 96.99 & 97.76 & 97.92 & 98.06\\
            \hline
        \end{tabular}
    \vspace{2mm}
    \caption{CS5 1:N Identification}
    \label{tab:cs5_identification}
\end{table*}

\begin{table*}
    \centering
        \begin{tabular}{|c|c|c|c|c|c|c|}
            \hline
            & \multicolumn{6}{|c|}{Retrieval Rate (\%) (Average for G1 and G2)}\\
            \hline
            \textbf{Method} & Rank=1 & Rank=2 & Rank=5 & Rank=10 & Rank=20 & Rank=50 \\
            \hline
            Fusion (Ours) & 97.18 & 97.65 & 97.90 & 98.04 & 98.16 & 98.31\\
            \hline
        \end{tabular}
    \vspace{2mm}
    \caption{CS5 1:N end-to-end still image identification}
    \label{tab:cs5_task9}
\end{table*}

%% file: dpssd_exp.tex
\subsection{Face Detection}
\label{sec:dpssd_results}

\begin{figure*}[htp!]
 \centering
\includegraphics[width=6.0cm]{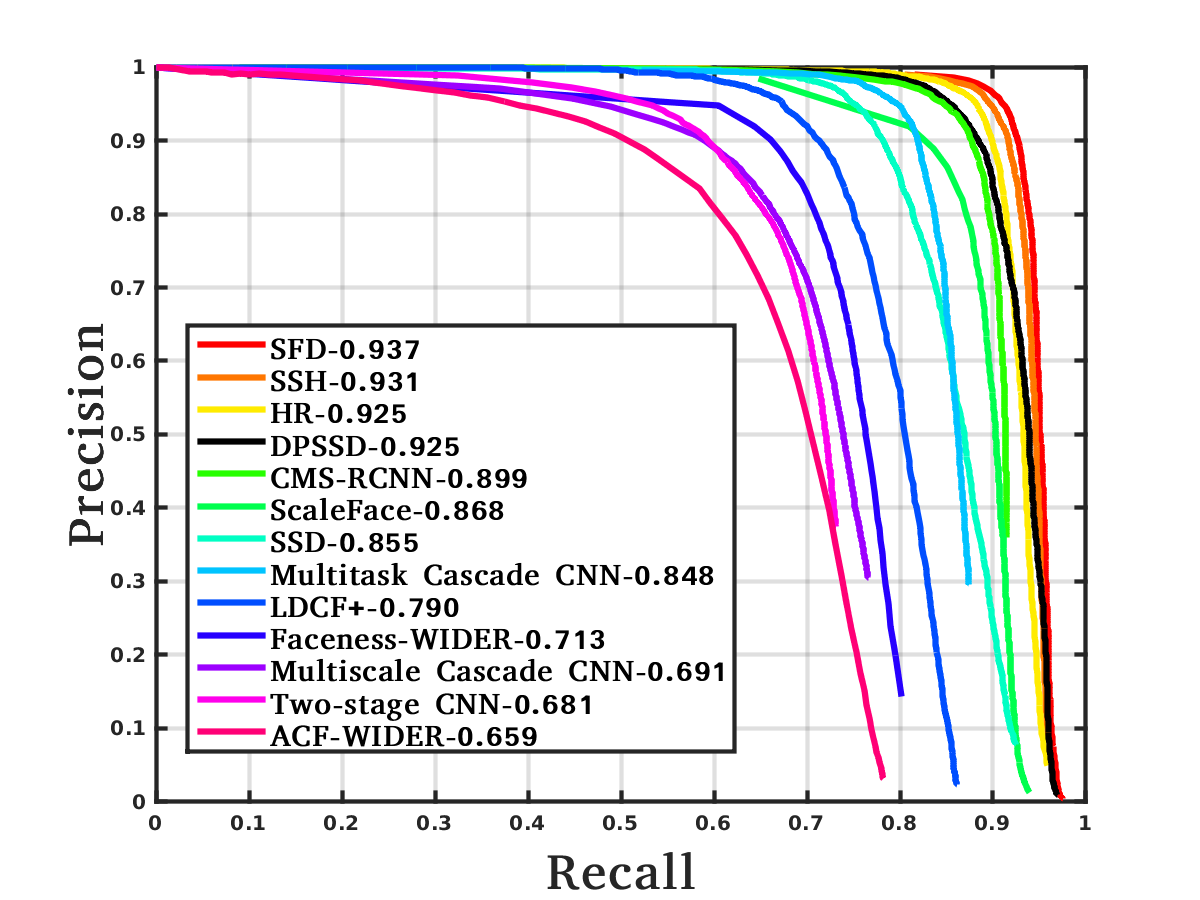}\hskip.1pt\includegraphics[width=6.0cm]{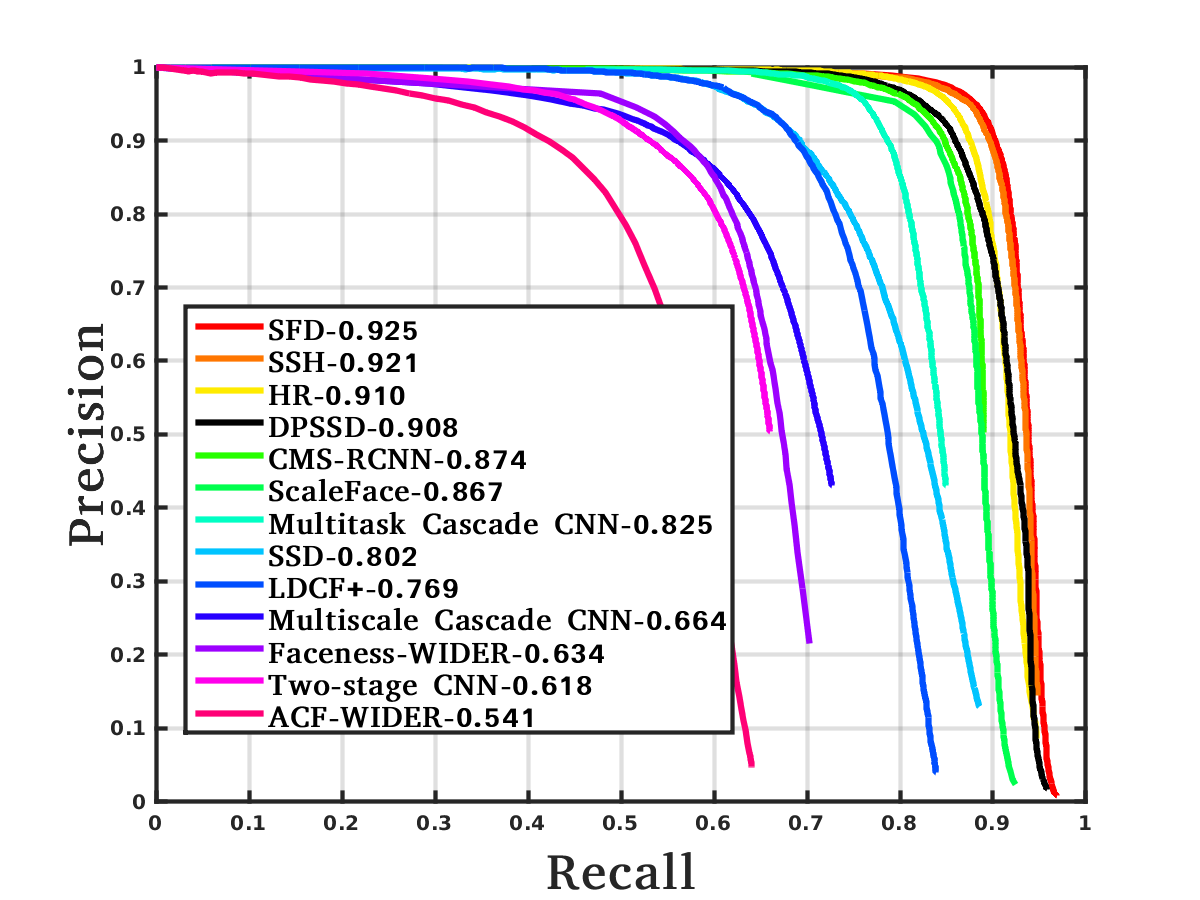}\hskip.1pt\includegraphics[width=6.0cm]{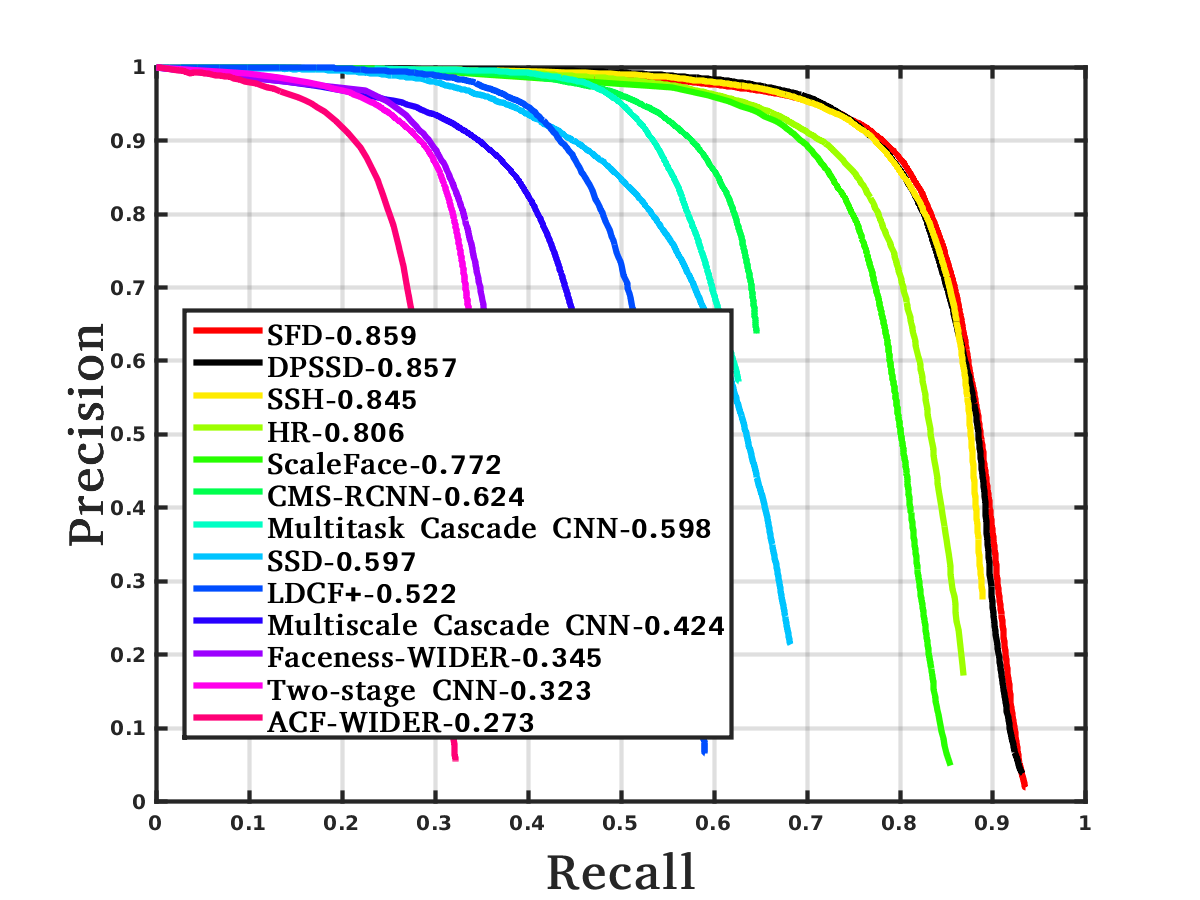}\\
(a) Easy \hskip130pt(b) Medium \hskip130pt(c) Hard\\
\caption{Performance evaluation on the WIDER Face~\cite{yang2016wider} validation dataset for (a)
    Easy, (b) Medium, and (c) Hard faces. The numbers in the legend represent the mean average
precision (mAP) for the corresponding method.}
\label{fig:WIDER_results}
\end{figure*}

We evaluated the proposed DPSSD face detector on four challenging face detection datasets: WIDER
Face~\cite{yang2016wider}, Unconstrained Face Detection Dataset (UFDD)~\cite{nada2018pushing}, Face
Detection Dataset and Benchmark (FDDB)~\cite{jain2010fddb} and Pascal Faces~\cite{PASCAL_faces}. We
achieve state-of-the-art performance on Pascal Faces~\cite{PASCAL_faces} dataset, and competing
results on WIDER~\cite{yang2016wider}, UFDD~\cite{nada2018pushing} and FDDB~\cite{jain2010fddb}
datasets.

\subsubsection{WIDER Face Dataset Results}
The dataset contains $32,203$ images with $393,703$ face annotations, out of which $40\%$ images are
used for training, $10\%$ for validation, and remaining $50\%$ for test. It contains rich
annotations, including occlusions, poses, event categories, and face bounding boxes. The faces
posses large variations in scale, pose and occlusion. The dataset is extremely challenging for the
task of tiny face detection, since the face width can be as low as $4$ pixels. We use the training
set to train the face detector and evaluate its performance on the validation set.
Fig.~\ref{fig:WIDER_results} provides the comparison of recently published face detection algorithms
with the proposed DPSSD.

We compare the performance of DPSSD with S$^{3}$FD~\cite{zhang2017s}, SSH~\cite{najibi2017ssh},
HR~\cite{hu2016finding}, CMS-RCNN~\cite{zhu2017cms}, ScaleFace~\cite{yang2017face}, Multitask
Cascade~\cite{zhang2016joint}, LDCF+~\cite{ohn2016boost}, Faceness~\cite{yang2015facial}, Multiscale
Cascade~\cite{yang2016wider}, Two-stage CNN~\cite{yang2016wider}, and ACF~\cite{yang2014aggregate}.
We observe that DPSSD achieves competing performance with state-of-the-art methods
(S$^{3}$FD~\cite{zhang2017s}, SSH~\cite{najibi2017ssh}, and HR~\cite{hu2016finding}). It achieves a
mean average precision (mAP) of $0.925$ and $0.908$ on easy and medium difficulty set, respectively.
On the hard set, it performs very close to the best performing method (S$^{3}$FD~\cite{zhang2017s})
with the mAP of $0.857$.

We also compare our method with the baseline face detector trained by fine-tuning
SSD~\cite{andy_ssd}. We outperform SSD~\cite{andy_ssd} on easy, medium as well as hard set.
Particularly on the hard set, DPSSD improves the mAP by a factor of $44\%$ over the
SDD~\cite{andy_ssd}. It shows that redesigning anchor pyramid with fixed aspect ratio, and adding
the upsampling layers helps tremendously in boosting the performance of face detection.

\subsubsection{UFDD Dataset Results}
\begin{figure}[htp!]
  \centering
  \includegraphics[width=5cm, height=4.5cm]{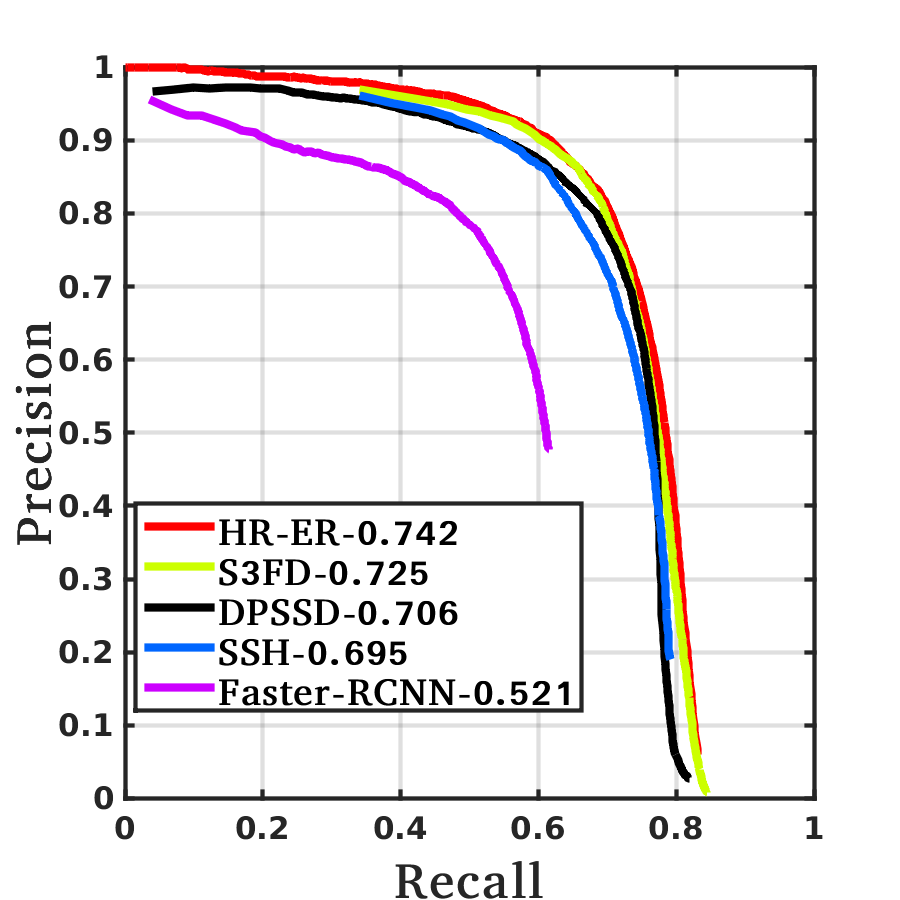}
  \caption{Performance evaluation on the UFDD~\cite{nada2018pushing} dataset. The numbers in the
  legend represent the mAP for the corresponding method.}
  \label{fig:ufdd_results}
  
\end{figure}

UFDD is a recent face detection dataset that captures several realistic issues not present in any
existing dataset. It contains face images with weather-based degradations (rain, snow and haze),
motion blur, focus blur, etc. Additionally, it contains distractor images that either contain
non-human faces such as animal faces or no faces at all, which makes this dataset extremely
challenging. It contains a total of $6,425$ images with $10,897$ face annotations. We compare our
proposed method with S$^{3}$FD~\cite{zhang2017s}, SSH~\cite{najibi2017ssh}, HR~\cite{hu2016finding},
and Faster-RCNN~\cite{jiang2016face} (see Fig.~\ref{fig:ufdd_results}). Similar to WIDER
Face~\cite{yang2016wider} dataset, we achieve competing results with mAP of $0.706$. Note that our
algorithm  is not fine-tuned on the UFDD dataset.

\subsubsection{FDDB Dataset Results}
\begin{figure}[htp!]
  \centering
  \includegraphics[width=0.45\textwidth]{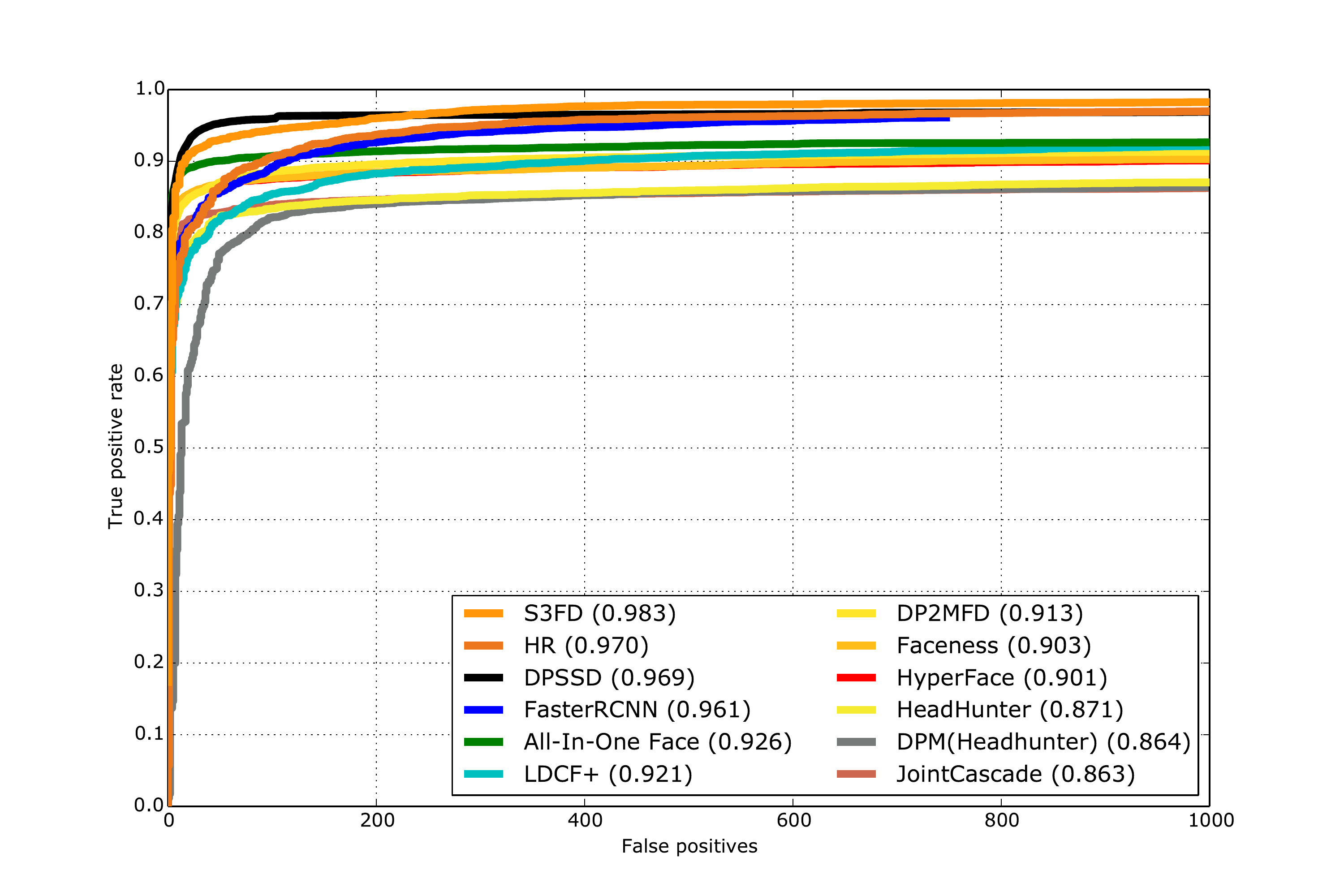}
  \caption{Performance evaluation on the FDDB~\cite{jain2010fddb} dataset. The numbers in the legend
  represent the mAP for the corresponding method.}
  \label{fig:fddb_results}  
\end{figure}

The FDDB dataset~\cite{jain2010fddb} is a benchmark for unconstrained face detection. It consists of
$2,845$ images containing a total of $5,171$ faces collected from news articles on the Yahoo
website. The images were manually localized for generating the ground truth. The dataset has two
evaluation protocols - discrete and continuous which essentially correspond to coarse match and
precise match between the detection and the ground truth, respectively. We evaluate the performance
of our method on the discrete protocol using the Receiver Operating Characteristic (ROC) curves, as
shown in Fig.~\ref{fig:fddb_results}.

We compare the performance of different face detectors such as S$^{3}$FD~\cite{zhang2017s},
HR~\cite{hu2016finding}, Faster-RCNN~\cite{jiang2016face}, All-In-One Face~\cite{ranjan2016all},
LDCF+~\cite{ohn2016boost}, DP2MFD~\cite{ranjan2015deep}, Faceness~\cite{yang2015facial},
HyperFace~\cite{ranjan2017hyperface}, Headhunter~\cite{HeadHunter_Mathias_ECCV2014},
DPM~\cite{HeadHunter_Mathias_ECCV2014}, and Joint Cascade~\cite{JointCascade_LI_ECCV2014}. As can be
seen from the figure, our method exhibits competing performance with state-of-the-art methods
(S$^{3}$FD~\cite{zhang2017s} and HR~\cite{hu2016finding}) and achieves a mAP of $0.969$. It should
be noted that our method does not use any fine-tuning or bounding box regression specific to the
FDDB dataset.

\subsubsection{PASCAL Faces Dataset Results}
\begin{figure}[htp!]
  \centering
  \includegraphics[width=0.4\textwidth]{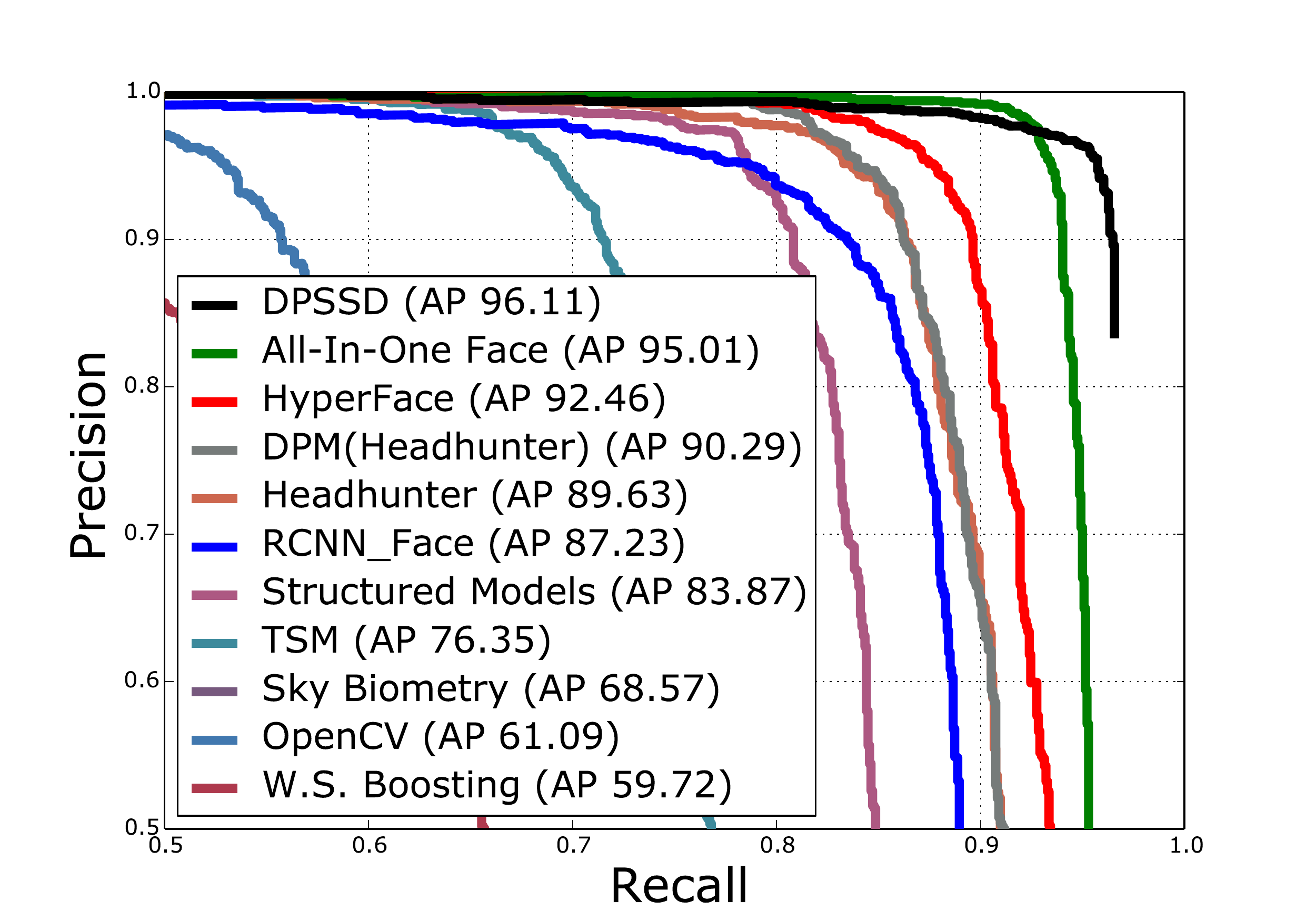}
  \caption{Performance evaluation on the Pascal Faces~\cite{PASCAL_faces} dataset. The numbers in
  the legend represent the mAP for the corresponding method.}
  \label{fig:pascal_results}  
\end{figure}

The PASCAL faces~\cite{PASCAL_faces} dataset was collected from the test set of the person layout
dataset which is a subset of PASCAL VOC~\cite{PASCALVOC}. The dataset contains $1,335$ faces from
$851$ images with large variations in appearance and pose. Fig.~\ref{fig:pascal_results} compares
the performance of different face detectors on this dataset. From the figure, we observe that our
proposed DPSSD face detector achieves the best mAP of $96.11\%$ on this dataset.

%% file: conclusion.tex
\section{Conclusions}
\label{sec:conclusion}

In this paper, we presented an overview of modern face recognition systems based on deep CNNs. We
discussed all parts of a face recognition pipeline and the state-of-the-art in those. We also
presented details of our face recognition system which uses an ensemble of two networks for
feature representation. Face detection and keypoint localization for in our pipeline is done using
the All-in-One CNN. We discussed training and datasets details for our system and how it connects to
existing work on face recognition. We presented results of our system for four challenging
datasets, \emph{viz.}, IJB-A, IJB-B, IJB-C, and IARPA Janus Challenge Set 5 (CS5). We show that our
ensemble based system achieves near state-of-the-art results. 

However, several issues remain unresolved even now. There is a need to develop theoretical understanding of
face recognition systems based on DCNNs. Given the multitude of loss functions used to train these
networks, we need to develop a unified framework which can put all of them in context to each other.
Domain adaptation and dataset bias is also an issue for current face recognition systems. These
systems are usually trained on a dataset and work well for similar test sets. However, networks
trained on one domain don't perform well for others. We trained our system on a combination of
different datasets. This made the trained models more robust. Training CNNs currently takes several
hours to days. There is a need to develop more efficient architectures and implementations of CNNs
which can be trained faster.

\section*{Acknowledgments}
This research is based upon work supported by the Office of the Director of National Intelligence
(ODNI), Intelligence Advanced Research Projects Activity (IARPA), via IARPA R\&D Contract No.
2014-14071600012. The views and conclusions contained herein are those of the authors and should not
be interpreted as necessarily representing the official policies or endorsements, either expressed
or implied, of the ODNI, IARPA, or the U.S. Government. The U.S. Government is authorized to
reproduce and distribute reprints for Governmental purposes notwithstanding any copyright annotation
thereon.